\pdfoutput=1

\documentclass{article}
\usepackage{iclr2027_conference,times}
\iclrfinalcopy

\usepackage[T1]{fontenc}
\usepackage{microtype}
\usepackage{amsmath,amssymb}
\usepackage{graphicx}
\usepackage[table]{xcolor}
\usepackage{booktabs}
\usepackage{multirow}
\usepackage{makecell}
\usepackage{tabularx}
\usepackage{array}
\usepackage{float}
\usepackage{placeins}
\usepackage{flafter}             
\usepackage{listings}
\usepackage{enumitem}
\usepackage{xspace}
\usepackage{etoolbox}

\definecolor{steel}{HTML}{2E4A73}
\definecolor{steel2}{HTML}{7F97BD}
\definecolor{jrdeep}{HTML}{1D5FA8}
\definecolor{keptrow}{HTML}{EEF2F8}
\definecolor{grayna}{gray}{0.6}

\usepackage[font=small,labelfont={bf,color=steel},labelsep=colon,skip=5pt]{caption}
\newcommand{\runninghead}{Preprint}
\makeatletter
\def\section{\@startsection{section}{1}{\z@}{-2.0ex plus -0.5ex minus -.2ex}{1.5ex plus 0.3ex minus 0.2ex}{\large\sc\raggedright\color{steel}}}
\def\subsection{\@startsection{subsection}{2}{\z@}{-1.8ex plus -0.5ex minus -.2ex}{0.8ex plus .2ex}{\normalsize\sc\raggedright\color{steel}}}
\renewenvironment{abstract}{\vskip.075in\centerline{\large\sc\color{steel}Abstract}\vspace{0.5ex}\begin{quote}}{\par\end{quote}\vskip 1ex}
\makeatother

\newcommand{\justrua}{JustRUA\xspace}
\newcommand{\na}{\textcolor{grayna}{--}}
\newcommand{\code}[1]{\texttt{#1}}
\newcommand{\kept}{\rowcolor{keptrow}}
\newcommand{\grouprow}[2]{\multicolumn{#1}{l}{\textit{#2}}\\}
\newcommand{\figpath}{figures/web}

\newcommand{\takeaway}[3]{%
  \par\addvspace{6pt}\noindent
  {\setlength{\fboxsep}{7pt}%
  \colorbox{keptrow}{\parbox{\dimexpr\linewidth-2\fboxsep\relax}{%
    \textbf{\textcolor{steel}{Takeaway #1:} #2}\enspace #3}}}%
  \par\addvspace{4pt}}

\setlist[itemize]{leftmargin=1.4em,itemsep=2pt,topsep=3pt}

\usepackage[colorlinks,linkcolor=jrdeep,citecolor=jrdeep,urlcolor=jrdeep,breaklinks]{hyperref}
\usepackage[capitalize,noabbrev]{cleveref}
\hypersetup{
  pdftitle={JustRUA: A Controlled Study of GPT-6 Robot-Use Agents},
  pdfauthor={Zefan Wang, Yufan Dang, Xuanwei Shi, Mingqi Gao, Cheng Yin, SikYuen Tam, Hanyu Liu, Yasheng Wang, Yaxi Lu, Songyang Zhang, Xin Cong, Yuan Yao},
  pdfsubject={Controlled study of harness, action interface, experience reuse and scene geometry for GPT-6 Astra robot-use agents on RoboDojo},
  pdfkeywords={robot-use agents, GPT-6 Astra, RoboDojo, harness, action interface, scene geometry, depth, self-improvement, bimanual manipulation}}
\crefname{listing}{Listing}{Listings}
\crefname{lstlisting}{Listing}{Listings}

\title{\textcolor{steel2}{Just}\textcolor{steel}{RUA}: A Controlled Study of\\GPT-6 Robot-Use Agents}

\author{%
\normalfont
\textbf{Zefan Wang}$^{1*\dagger}$\quad
\textbf{Yufan Dang}$^{1*}$\quad
\textbf{Xuanwei Shi}$^{1*}$\quad
\textbf{Mingqi Gao}$^{2*}$\\
\normalfont
\textbf{Cheng Yin}$^{3}$\quad
\textbf{SikYuen Tam}$^{1}$\quad
\textbf{Hanyu Liu}$^{4}$\quad
\textbf{Yasheng Wang}$^{2}$\\
\normalfont
\textbf{Yaxi Lu}$^{1}$\quad
\textbf{Songyang Zhang}$^{5}$\quad
\textbf{Xin Cong}$^{1}$\quad
\textbf{Yuan Yao}$^{1\ddagger}$\\[5pt]
\normalfont\small
$^{1}$Tsinghua University\quad
$^{2}$ModelBest\quad
$^{3}$Huazhong University of Science and Technology\\
\normalfont\small
$^{4}$Peking University\quad
$^{5}$Tencent\\[2pt]
\normalfont\small
$^{*}$Core contributor\quad
$^{\dagger}$Project lead\quad
$^{\ddagger}$Corresponding author\\[6pt]
\normalfont\small
\href{https://zefanw.github.io/JustRUA}{Project page}\enspace\textcolor{black!35}{|}\enspace
\href{https://github.com/ZefanW/JustRUA}{Code}\enspace\textcolor{black!35}{|}\enspace
\href{https://huggingface.co/spaces/ZefanW/robodojo-replay-explorer}{Replays}\enspace\textcolor{black!35}{|}\enspace
\texttt{zefan.wang.thu@gmail.com}\enspace\textcolor{black!35}{|}\enspace
\texttt{yaoyuanthu@gmail.com}%
}

\begin{document}

\maketitle

\begin{abstract}
GPT-6 Astra can drive a robot arm directly through a fixed harness, and scores of such robot-use agents on RoboDojo have risen quickly since.
Reported systems differ in harness and test subset and often change several designs at once, so it is hard to tell which choices produce the gains.
We take these designs apart and test them in one framework, along the harness around the model, the action interface, the reuse of experience and the scene geometry given to the model.
A lean harness that keeps the full episode history and stops each motion early goes a long way.
Reusing experience changes how the model acts but is not a free lunch, and scene geometry helps most, even when depth is predicted from RGB alone.
\justrua starts from the official L3 harness and keeps only what matters most.
Even with no learned policy, self-improvement loop or task-specific tuning, this simple harness raises the score on the RoboDojo simulation benchmark substantially, from 29.7 to 50.3.
\end{abstract}

\begin{figure}[H]
\centering
\includegraphics[width=0.98\linewidth]{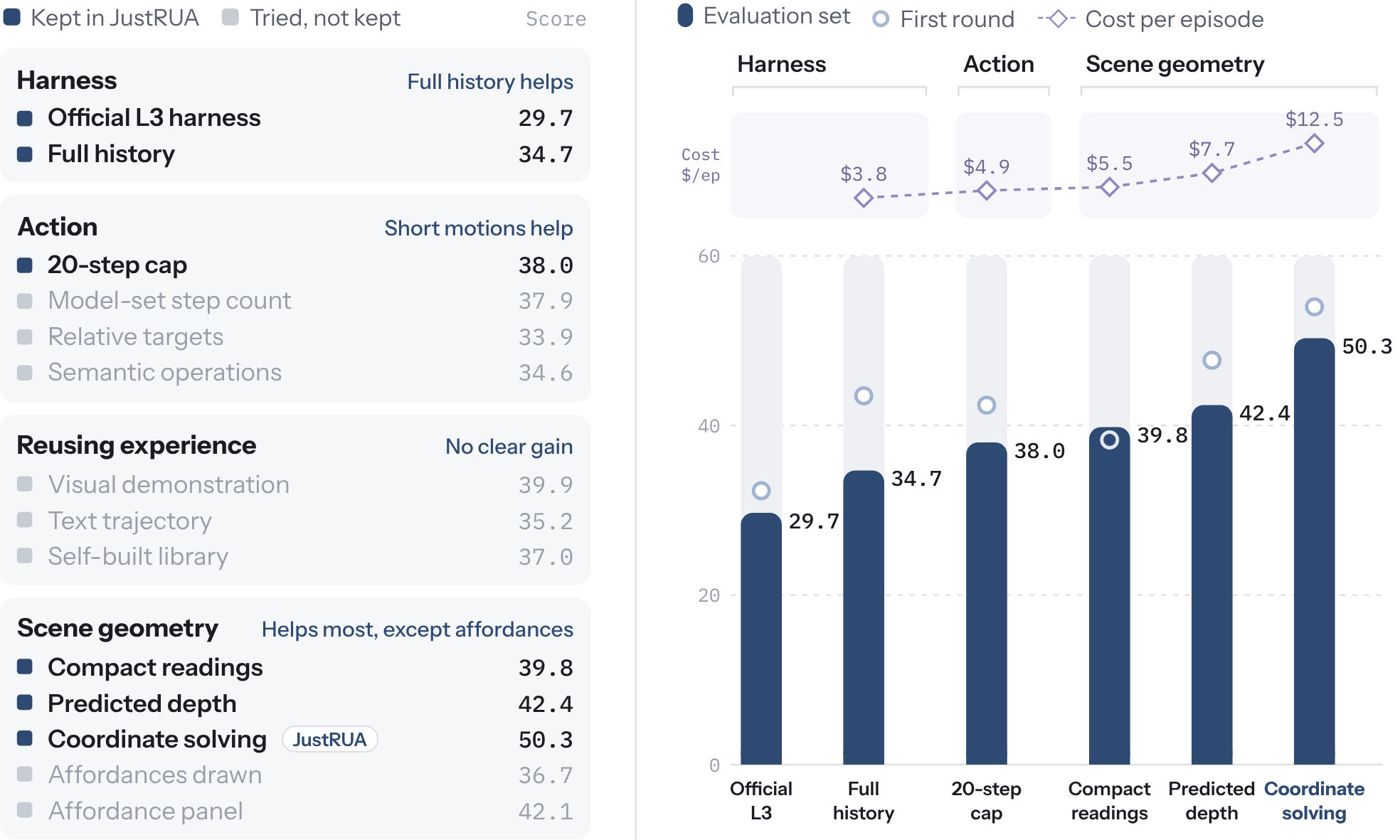}
\caption{\textbf{From the official L3 harness to \justrua.}
Left, every design we tested, by axis, with its score on the evaluation set, 111 episodes that cover all 42 tasks.
Right, the designs \justrua keeps, in the order they were added.
Each bar is one evaluation-set run, the ring its first-round subset of 42 episodes, and the dashed line the cost of one first-round episode, a lower bound that assumes all cacheable input hits the cache.
Reusing experience draws on the tasks being scored, so none of its forms is kept, whatever the score.}
\label{fig:teaser}
\end{figure}

\clearpage

\section{Introduction}
\label{sec:intro}

Frontier multimodal models can now control robots without any robot-specific training.
GPT-6 Astra \citep{openai2026gpt6astra} drives a bimanual arm through a harness, the software that turns each of its decisions into end-effector targets for a motion planner \citep{zhang2026roboprobe,galbot2026exploring,yang2026astranomy}.
No vision-language-action (VLA) model sits in between.
The model looks at camera images, decides where each gripper should go next and reads the outcome, much as a computer-use agent looks at a screen and decides where to click \citep{xie2024osworld}.
By analogy such a system is called a robot-use agent (RUA) \citep{isola2026rua,yang2026ora}.
On the RoboDojo simulation benchmark \citep{chen2026robodojo}, RUAs built on GPT-6 Astra have improved quickly, and each new system adds components of its own, such as depth and point clouds \citep{qu2026roboharness,zhou2026gptgear,huang2026rax,wang2026auriga}, affordance models \citep{kim2026claro,yang2026ora}, in-context demonstrations \citep{guo2026robodawn,cai2026sensenova,liu2026roboicl} and self-improvement loops that write and refine their own skills \citep{li2026embodiedevo,sensenova2026roborsi}.

Our starting point was a discrepancy.
The first evaluation of GPT-6 Astra on RoboDojo found it weak on tasks that require precision under the official L3 harness \citep{zhang2026roboprobe}, and a broader study saw a similar gap between its task decisions and reliable physical control \citep{galbot2026exploring}.
Yet a few reasonable fixes to the harness and the action loop, with no elaborate design on top, brought its Precision score to the level of the strongest VLA models of the time (\cref{sec:harness}).
We came to suspect that much of the reported weakness lay in the harness rather than in the model.

If the harness matters this much, the gains reported by recent systems are hard to read.
These systems start from different harnesses, run on different subsets of the benchmark with different budgets, and often change several parts of the design at once.
A reader who sees a higher score cannot tell whether it comes from the new component, from the loop it was added to or from the episodes it was measured on.
Reproducing each system in full would carry these differences over.
We instead take the designs apart and test them on one testbed and one evaluation set, changing as little as we can between runs.

\begin{figure}[t]
\centering
\includegraphics[width=\linewidth]{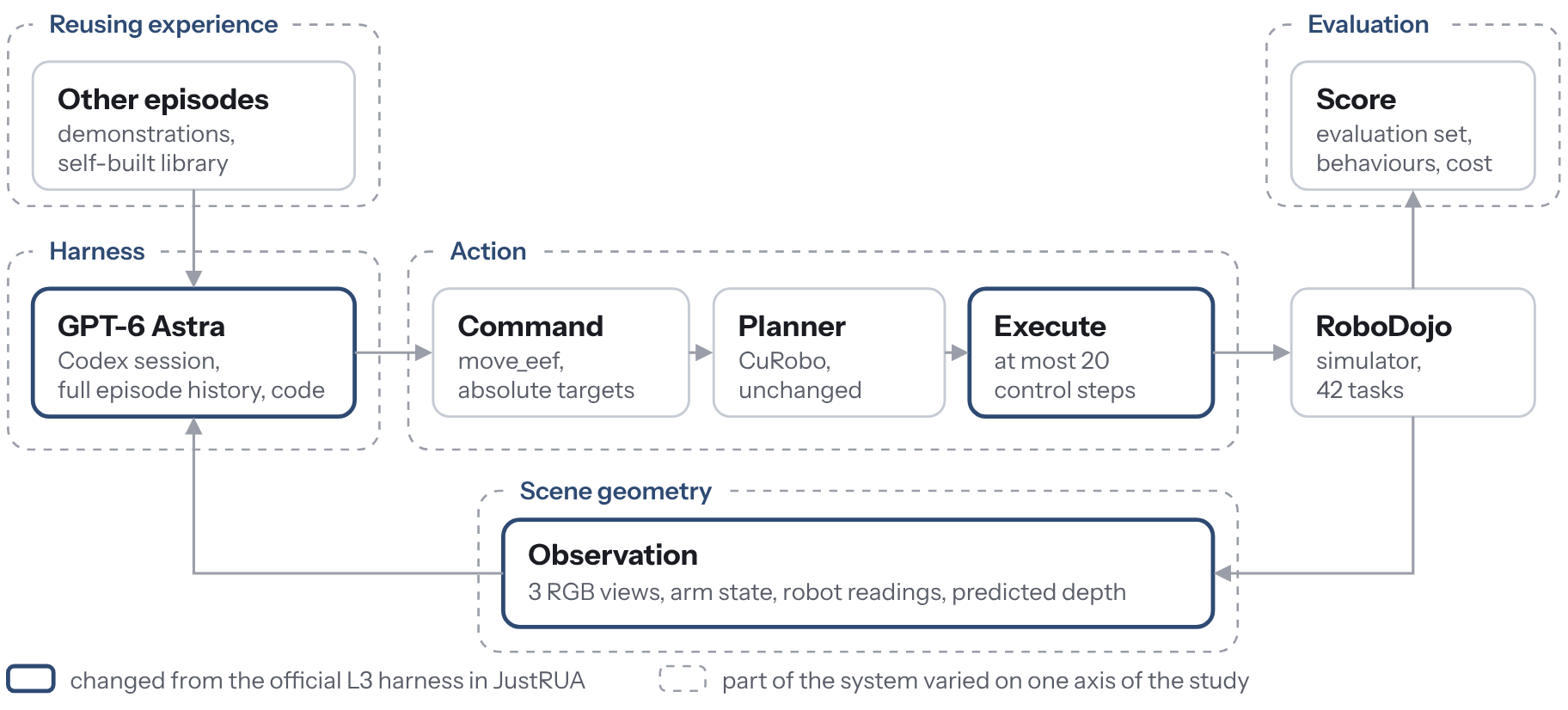}
\caption{\textbf{The testbed loop and the four axes of the study.}
The model issues a command, the unchanged planner turns it into a motion, and the simulator returns the next observation and, at the end, the score.
Solid blue outlines mark the parts \justrua changes from the official L3 harness, and each dashed area is one axis of the study.
\Cref{tab:axes} in the appendix describes every variant on each axis.}
\label{fig:framework}
\end{figure}

We start from the official L3 harness and group the designs that recent systems add along four axes (\cref{fig:framework}): the harness around the model, the action interface, the reuse of experience from other episodes and the scene geometry in each observation.
Three tools keep the comparison affordable and readable.
Since the official evaluation of 2,100 episodes is costly to repeat for every design, we build a 111-episode evaluation set that covers all 42 tasks and spends more episodes on the tasks with high evaluation uncertainty.
Since a score says how much of a task was completed and not how the attempt went, we review every first-round episode from its video and count behaviours the score does not show, such as empty grasps and dropped objects.
And since an episode resends its whole history at every decision, we report what each configuration costs per episode.
A design is kept only if it helps on the evaluation set at a reasonable cost and does not draw on the tasks being scored.

The result is \justrua.
It keeps three kinds of change to the official harness.
The loop keeps the full episode history and stops each motion after 20 control steps, the observation carries the robot's own grip readings and depth predicted from RGB, and the model may run code to solve 3D coordinates from pixels.
With no learned policy, no self-improvement loop and no task-specific tuning, it raises the evaluation-set score of the official harness from 29.7 to 50.3 (\cref{fig:teaser}).
Our findings are as follows.

\begin{itemize}
\item \textbf{A lean harness goes a long way} (\cref{sec:harness}).
Keeping the full episode history and stopping each motion after 20 control steps lift the score from 29.7 to 38.0.
Relative targets, named skills and model-chosen step counts do not beat plain absolute targets.
\item \textbf{Reusing experience is not a free lunch} (\cref{sec:experience}).
Demonstrations and a self-written skill library change how the model acts, but neither clearly beats the default, and both make an episode more expensive.
They also draw on the tasks being scored, so their scores are not comparable with a standard evaluation.
\item \textbf{Scene geometry helps most} (\cref{sec:geometry}).
Depth images cut empty grasps from 41 to 16 in the first round, even when Depth Pro predicts the depth from RGB alone and is off by tens of centimetres.
With code that solves 3D coordinates from pixels and no decision cap, the score reaches 50.3.
\item \textbf{Remaining failures are mostly long-tail} (\cref{sec:errors}).
Most failures left in our best configuration are specific to one task, such as plans that do not suit the arm, unplanned knocks and slipping grasps.
Fixing them would take trial and error on each task, which comes close to tuning on the test tasks.
\end{itemize}

The evaluation set, the code and the episode replays are available from the project page.

\section{Controlled Study Framework}
\label{sec:framework}

Instead of reproducing each reported system, we take the designs apart and test them on one testbed and one benchmark, changing as little as we can between runs.
The framework has to meet two conditions.
It must carry out the actions the model chooses reliably, so that we measure the model and its interface rather than the accidents of the loop around them.
And it must tell the effects of different designs apart at a cost we can pay many times over.
This section describes the benchmark and the harness we start from, the testbed, the four axes, the evaluation set, and the review of videos and logs that we use to count physical behaviours.

\subsection{Benchmark and Starting Point}
\label{sec:benchmark}

RoboDojo \citep{chen2026robodojo} follows earlier bimanual simulation benchmarks such as RoboTwin 2.0 \citep{chen2025robotwin2} and contains 42 simulated bimanual tasks in five capability groups, Generalization, Precision, Long-horizon, Memory and Open.
Each task is scored from 0 to 100, where full success scores 100 and many tasks give partial credit for intermediate stages.
The official evaluation runs 50 episodes per task on different layouts, 2,100 episodes in all, and averages tasks within each group and then the five groups with equal weight.

Our starting point is L3 Inspect EEF, the official L3 harness released with RoboProbe \citep{zhang2026roboprobe} and the one behind the GPT-6 Astra entry on the RoboDojo leaderboard.
RoboProbe sorts model-driven robot systems by what sits in the action path, and at its third level (L3) the model acts through a fixed, non-learned harness with no learned policy in between.
Each episode opens with the task instruction and a task recipe from the RoboDojo wiki, which describes the task and its scoring stages without giving an action sequence.
At each decision the model sees three RGB views, from the head and both wrists, and the measured grasp-point and joint state of both arms.
It acts by calling \code{move\_eef} with world-frame targets for the grasp point of either arm, given as position, orientation and gripper opening, or \code{give\_up} to end an episode it judges unrecoverable.
Non-learned code plans a joint path to each target with CuRobo \citep{sundaralingam2023curobo}, the simulator plays it back, and the next observation reports how far the grasp point ended up from its target.
The model decides what to do next and where to move, and the planner decides how the arm gets there.
The model receives no depth, object poses, layout metadata or scoring signal, and the simulator alone decides whether an episode succeeds.
The official harness keeps images from the last two observations only, and an episode ends at the latest when the benchmark's limit of 500 control steps runs out.

\subsection{Testbed}
\label{sec:testbed}

We run the same loop in a Codex \citep{openai2026codex} session at medium reasoning effort.
Codex already manages a long session and compacts the context as it nears the window limit, so we did not rebuild these parts ourselves.
We use it only to hold the conversation.
Its own tools, such as the shell, web search and its image viewer, are switched off.
The definitions of \code{move\_eef} and \code{give\_up} are given to the model as text in a fixed instruction, the model returns each decision as one JSON object that holds its note and its calls, and the unchanged code of the official harness parses the calls, plans each motion and executes it.
The planner and its inverse kinematics are those of the official harness.

Within one episode the session appends every observation, action and feedback message, images included, and nothing carries over from one episode to the next.
Every experiment runs on Codex 0.153.4, pinned so that small differences between versions cannot move the results.
An episode has three separate limits, the number of model decisions, the control steps executed per motion and the benchmark's own limit of 500 control steps.
All controlled comparisons allow 100 decisions per episode, and only the final configuration of \justrua lifts this limit (\cref{sec:coord}).
Apart from the task recipes inherited from the official harness, no configuration contains a task-specific tool, prompt or piece of code (\cref{app:testbed}).

\subsection{Four Axes}
\label{sec:axes}

We group the designs that recent systems add along four axes (\cref{fig:framework}).
The \emph{harness} axis covers the loop around the model, that is, how much of the episode it keeps and how hard the model reasons.
The \emph{action} axis covers how a decision becomes motion, namely the command format and how long each motion runs before the model looks again.
The \emph{reusing experience} axis covers what the model carries over from other episodes, ordered by how far the experience is condensed, from a visual demonstration to a text trajectory and a library of operations the system writes for itself.
The \emph{scene geometry} axis covers what each observation says about the 3D layout of the scene, from the robot's own readings to depth images, precomputed affordances and coordinates the model solves in code.

We move along the axes in this order and keep a design when it helps on the evaluation set at a reasonable cost and does not draw on the tasks being scored.
Each alternative is therefore compared with the configuration kept at the point where it was tested, and the kept designs form a single path from the official harness to \justrua (\cref{fig:teaser}).
This is a path and not a full factorial design.
It tells us whether a design helps on top of what is already kept, which is the question a system builder faces, and it does not test every combination.

\subsection{Evaluation Set}
\label{sec:evalset}

Running 2,100 episodes for every design is costly, since one episode reads several million input tokens.
Running one episode per task is cheap but unreliable, because the score of a task on a single layout is spiky, a known hazard of robot policy evaluation \citep{kressgazit2024robotlearning}.
To find where the swings come from, we ran the default testbed (\cref{sec:harness}) for eight rounds at both low and medium reasoning effort, each round taking the next layout of every task in a fixed order.
Most of the swings came from one kind of task, where success scores 100, any partial result scores at most 50, and the model succeeds on some layouts but not on others, so that a single near-miss moves the task by 50 points or more.
We say such a task has high evaluation uncertainty.

\begin{figure}[!htb]
\centering
\includegraphics[width=\linewidth]{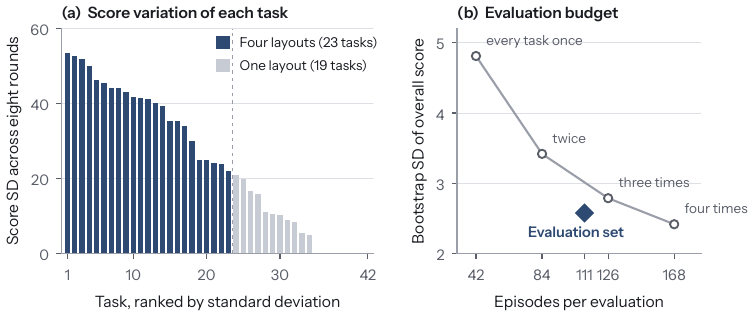}
\caption{\textbf{Evaluation set.}
(a) Standard deviation of each task's score across eight rounds of the default testbed.
The 23 tasks that run on four layouts are the 23 that vary most.
(b) Bootstrap standard deviation of the overall score at medium effort when every task runs once to four times, and for the evaluation set.
With 111 episodes the evaluation set varies less than running every task three times with 126.}
\label{fig:evalset}
\end{figure}

The evaluation set keeps all 42 tasks and spends repeated episodes where a single one is least reliable.
A task runs on four fixed layouts if a partial result scores at most 50 and the default testbed both succeeded and failed on it within the eight rounds.
This selects 23 tasks, which are exactly the 23 with the highest variance across rounds (\cref{fig:evalset}a).
The other 19 tasks run on one layout, for 111 episodes in total.
Layouts differ noticeably in difficulty, so they are fixed in advance and shared by every configuration, and repeated tasks gain no extra weight.
Scores are averaged over the layouts of a task, then over the tasks of a group, then over the five groups, as in the official score.

A bootstrap over the eight rounds gives the standard deviation of the overall score for each way of spending the budget (\cref{fig:evalset}b).
The evaluation set varies less than running every task three times, with fewer episodes.
For the official harness its score also stays close to that of the full evaluation (\cref{tab:main}).
The evaluation set is a tool for comparing designs under fixed conditions, not a substitute for the full evaluation, and \cref{app:eval} lists its tasks and its known differences from the official protocol.

\subsection{Behaviour Review and Cost}
\label{sec:review}

\paragraph{First round.}
The first round runs the first layout of each task, 42 episodes, and is a subset of the evaluation set.
A score from the first round alone tends to run high (\cref{app:eval}), so we compare configurations on the evaluation set and use the first round to study behaviour and cost.

\begin{figure}[!htb]
\centering
\includegraphics[width=\linewidth]{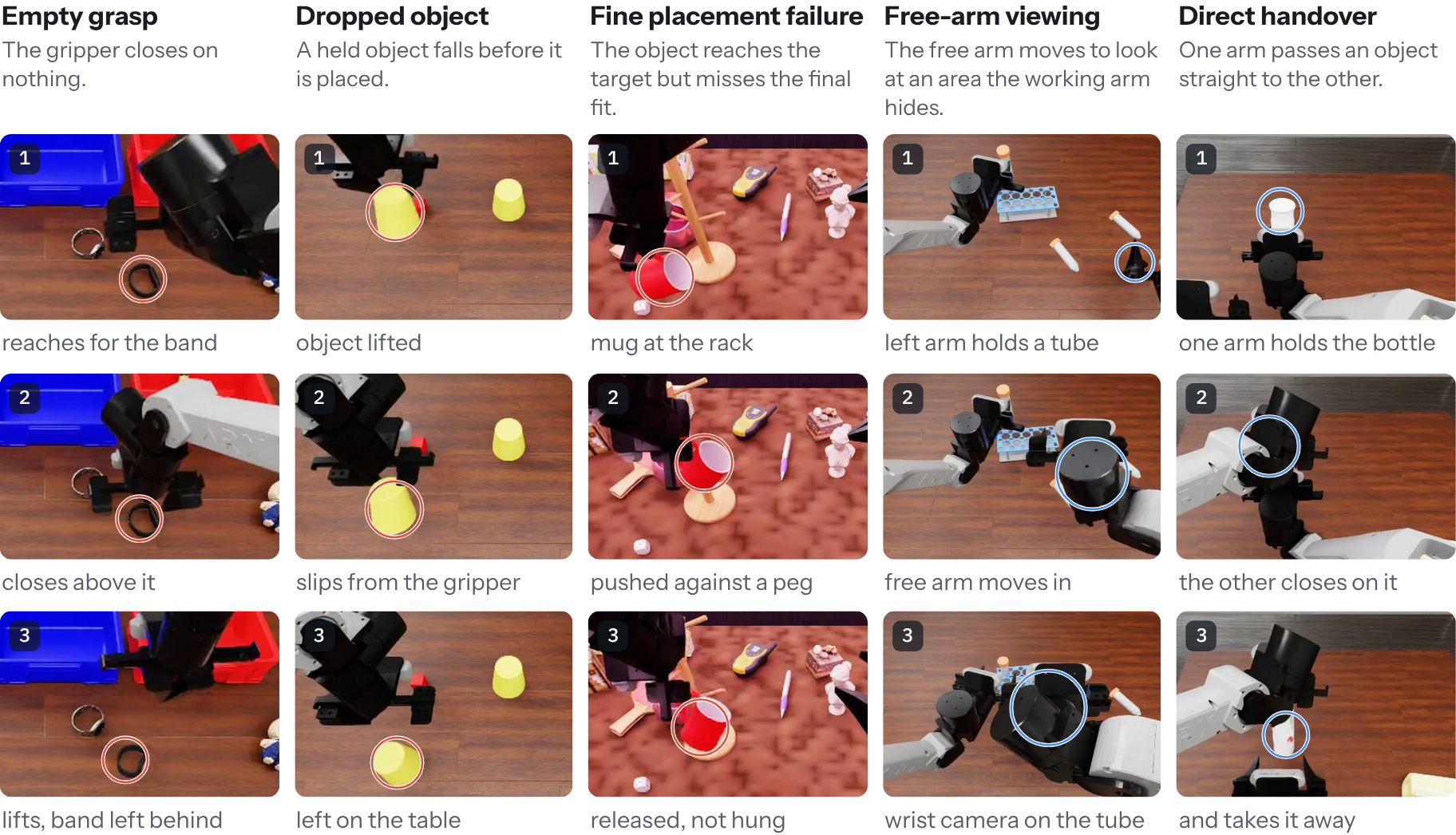}
\caption{\textbf{Behaviours counted in the review}, head camera, three moments of one episode each, from top to bottom.
Rings mark the object, or in the fourth column the free arm.
In the first column the gripper closes above a watch band and comes up empty.
The first three behaviours are failures and the last two are neutral.
Handovers through the table and unwanted contact are counted in the same way (\cref{app:review}).}
\label{fig:behaviours}
\end{figure}

\paragraph{Behaviour review.}
A score says how much of a task was completed, not how the attempt went.
Two configurations with the same score can fail in different ways, one by closing the gripper on nothing and another by dropping objects on the way.
For each configuration we therefore review the 42 first-round episodes and count a small set of physical behaviours (\cref{fig:behaviours}).
Each episode is reviewed on its own by a separate model, GPT-6.1 Sol, which takes no part in running the robot, from the full action log and the videos of all three cameras.
The reviewer scans each video at short intervals and looks closely around every executed action, and each counted event is tied to the frames that show it.
A second pass checks every episode's counts against this evidence and the native logs.
Only events the evidence settles are counted, so the counts are lower bounds, and we compare them only across configurations reviewed on the same episodes with the same written definitions (\cref{app:review}).
With one episode per task, the counts show a direction rather than settle small differences.

\paragraph{Cost.}
We report what an episode costs rather than how long it takes.
Time can be hidden by running episodes in parallel, and at this stage we are measuring what the system can do, not deploying it.
Cost cannot be hidden.
Because the full history is resent at every decision, an episode reads several million input tokens and writes only a few thousand, and nearly every input token has already been sent earlier in the same episode.
We therefore report the cost of one first-round episode with all such tokens billed at the cached-input price.
This lower bound does not depend on how well the provider's cache happened to work during a run.
Token use has also become a design target of its own for GPT-6 Astra agents \citep{si2026fewertokens}.

\section{Harness and Action}
\label{sec:harness}

\begin{table}[t]
\centering
\caption{\textbf{Every configuration on the evaluation set.}
Scores by task group and overall, success rate, and cost per first-round episode.
Each block states the configuration it builds on, and each affordance variant is compared with depth from the same source.
Simulator depth is not accepted by the RoboDojo leaderboard for agent entries and serves as a reference.
Shaded rows are kept in \justrua.
$^\dagger$Group scores and success rate of the official harness are from its full evaluation of 2,100 episodes, where it scores 29.0 overall.}
\label{tab:main}
\small
\setlength{\tabcolsep}{4.2pt}
\begin{tabular}{l ccccc cc c}
\toprule
Configuration & Gen. & Prec. & Long. & Mem. & Open & Overall & \makecell{Success\\(\%)} & \makecell{Cost\\(\$)} \\
\midrule
\grouprow{9}{Harness}
\kept Official L3 harness$^\dagger$ & \textcolor{grayna}{33.4} & \textcolor{grayna}{12.7} & \textcolor{grayna}{21.5} & \textcolor{grayna}{43.0} & \textcolor{grayna}{34.4} & 29.7 & \textcolor{grayna}{22.5} & \na \\
\kept Codex, full history & 36.0 & 9.4 & 24.5 & 52.5 & 50.9 & 34.7 & 27.1 & 3.76 \\
\midrule
\grouprow{9}{Action, on full history}
\kept 20-step cap (default testbed) & 42.5 & 25.0 & 24.7 & 49.6 & 48.1 & 38.0 & 31.0 & 4.94 \\
Model-chosen step count & 40.4 & 14.4 & 32.5 & 60.4 & 41.6 & 37.9 & 30.0 & 4.95 \\
Relative targets & 27.1 & 10.9 & 28.0 & 56.7 & 46.9 & 33.9 & 27.1 & 4.60 \\
Semantic operations & 23.3 & 21.9 & 28.8 & 61.0 & 38.1 & 34.6 & 29.2 & 5.97 \\
Move sequences & 44.0 & 20.0 & 40.5 & 58.3 & 47.2 & 42.0 & 30.8 & \na \\
\midrule
\grouprow{9}{Reasoning effort, on the default testbed (medium)}
Low effort & 40.3 & 5.3 & 28.8 & 52.9 & 39.4 & 33.3 & 27.3 & 4.49 \\
High effort & 38.3 & 17.2 & 23.8 & 67.3 & 40.6 & 37.4 & 30.6 & 5.16 \\
\midrule
\grouprow{9}{Reusing experience, on the default testbed}
Visual demonstration & 41.0 & 22.8 & 28.9 & 70.0 & 36.6 & 39.9 & 35.8 & 5.11 \\
Text trajectory & 40.3 & 11.6 & 28.3 & 55.4 & 40.3 & 35.2 & 29.0 & 5.84 \\
Self-built library & 35.2 & 15.9 & 37.3 & 58.3 & 38.1 & 37.0 & 26.3 & 5.43 \\
\midrule
\grouprow{9}{Scene geometry, each block on the kept row before it}
\kept Compact robot readings & 40.8 & 19.7 & 27.5 & 64.6 & 46.6 & 39.8 & 33.3 & 5.51 \\
Simulator depth (reference) & 42.4 & 28.8 & 29.2 & 72.9 & 56.9 & 46.0 & 38.3 & 7.26 \\
\kept Predicted depth & 52.3 & 20.6 & 31.6 & 55.6 & 51.9 & 42.4 & 35.4 & 7.74 \\
Affordances drawn, simulator depth & 37.4 & 9.4 & 21.7 & 72.1 & 43.1 & 36.7 & 30.8 & 6.91 \\
Affordances drawn with depth, predicted & 35.0 & 6.9 & 23.8 & 42.3 & 46.6 & 30.9 & 24.8 & 6.63 \\
Affordance panel, predicted depth & 38.8 & 26.6 & 26.1 & 69.2 & 50.0 & 42.1 & 34.8 & 7.54 \\
\kept Coordinate solving (\justrua) & 47.3 & 29.1 & 52.2 & 70.8 & 52.2 & \textbf{50.3} & \textbf{43.3} & 12.53 \\
\bottomrule
\end{tabular}
\end{table}

The first two axes concern the loop around the model and the form of its commands.
They are where the discrepancy of \cref{sec:intro} comes from.
With the two changes of this section, the Precision score of GPT-6 Astra on our evaluation set comes close to that of DM0.5 \citep{dexmal2026dm05}, the strongest VLA on the RoboDojo leaderboard at the time \citep{robodojo2026leaderboard}.
\Cref{tab:main} lists every configuration of the study, and this section covers its first three blocks.

\subsection{Codex and Full History}
\label{sec:history}

The official harness calls the model API directly and keeps images from the last two observations only.
Earlier observation text, notes, calls and results stay in the context, while earlier images drop out.
Our testbed runs the episode in one Codex session that keeps every image.
The model can then draw on what it has already seen and tried, such as an object that is now hidden behind the arm or the layout just before a contact.
This history covers the current episode only, so it is not a form of experience reuse.
In the default testbed the history also stays complete, since Codex never had to compact it in the episodes we inspected.

With the full history the evaluation-set score rises (\cref{tab:main}).
Moving the loop into Codex without the full history did not help by itself, so we credit the gain to the two together.

\subsection{Motion Length}
\label{sec:motion}

\begin{figure}[!htb]
\centering
\includegraphics[width=0.94\linewidth]{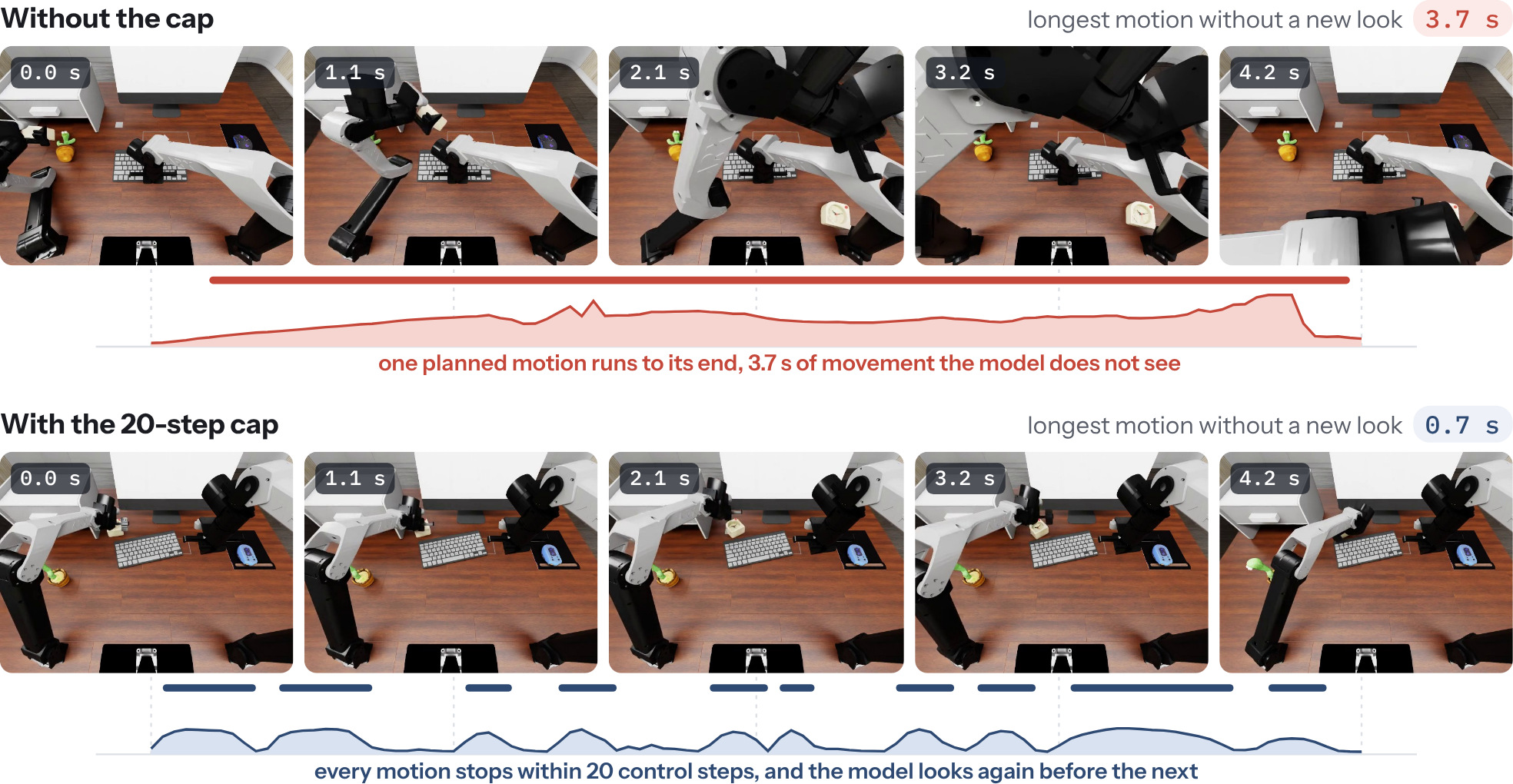}
\caption{\textbf{Motion length.}
The same 4.2 seconds of \code{organize\_table} without the cap and with it, five head-camera frames each.
The trace below the frames is the movement in the head view, and each bar marks a stretch of continuous movement.
Without the cap, one unstable plan carries the arm across the table and in front of the camera before the model sees anything.
With the cap, every motion stops within 20 control steps, about 0.8\,s.}
\label{fig:motion}
\end{figure}

Watching the episodes, we found that CuRobo occasionally returns an unstable plan, after which the arm makes a large and clearly unreasonable motion.
In the official loop the whole motion runs before the model sees anything, so such a motion completes unchecked (\cref{fig:motion}).
We keep the planner and its timing but execute at most the first 20 control steps of each plan, about 0.8 seconds of motion.
The trajectory is not resampled into 20 steps.
The rest of the plan, including any gripper command in it, is discarded rather than continued in the background, and the gripper holds the last command that was actually executed.
The model then receives new images and the measured state.
The feedback is still computed against the original target and states how many of the planned steps ran, so the model learns that the arm has not arrived yet instead of being told that the move is complete.
It may then keep approaching the same target, or change the pose, the gripper or the plan based on what it now sees.

The cap does not make the planner more accurate.
It shortens the stretch the arm moves without a new look, which limits how far a bad motion can go and gives the model an earlier chance to correct it.
The score rises again, and the gain concentrates in the Precision group (\cref{tab:main}).
Shorter motions mean more decisions, so an episode also costs more.

We also let the model choose the length of each motion itself, between 3 and 32 control steps, as in action chunking \citep{zhao2023act,yang2026astranomy}.
Its choices split between short and long motions, and it reaches the same overall score with a different balance across groups, higher on Long-horizon and Memory tasks and lower on Precision.
Since the extra freedom buys nothing overall, \justrua keeps the fixed cap as the simpler choice.
Full history with the 20-step cap at medium effort is the \emph{default testbed} for the rest of the study.

\subsection{Alternative Action Interfaces}
\label{sec:interfaces}

Two alternatives to absolute targets come from recent systems, and neither beats them (\cref{tab:main,tab:action-behaviour}).
\Cref{app:interfaces} defines each interface and shows one call of each kind at the same moment of a task.

\begin{table}[!htb]
\centering
\caption{\textbf{Behaviour under each action interface}, counted over the same 42 first-round episodes.
Absolute targets with the 20-step cap are the default testbed.}
\label{tab:action-behaviour}
\small
\setlength{\tabcolsep}{6pt}
\begin{tabular}{@{}lccccc@{}}
\toprule
& \makecell{Absolute\\targets} & \makecell{Relative\\targets} & \makecell{Semantic\\operations} & \makecell{Move\\sequences} & \makecell{Model-chosen\\step count} \\
\midrule
Empty grasps & 47 & 54 & 47 & 38 & 45 \\
Dropped objects & 26 & 30 & 35 & 27 & 29 \\
Fine placement failures & 20 & 25 & 19 & 20 & 22 \\
View adjustments & 59 & 54 & 211 & 45 & 61 \\
Handovers & 15 & 15 & 18 & 14 & 14 \\
Arm hits an object & 7 & 20 & 12 & 1 & 11 \\
\bottomrule
\end{tabular}
\end{table}

\paragraph{Relative targets.}
Relative targets give position and orientation as offsets from the pose measured at the current decision, as in RoboDawn \citep{guo2026robodawn}.
They score lower, much as Astranomy found \citep{yang2026astranomy}, and the arm strikes objects almost three times as often.
One likely reason is that every coordinate the model reads, from the task recipe to the measured pose, is given in the world frame, so an offset asks for one more subtraction before each move.

\paragraph{Semantic operations.}
Semantic operations replace raw targets with named skills \citep{ahn2022saycan,liang2023cap,chen2026showharness}.
Ours are built on \code{move\_eef} and only describe moves at a higher level, with no new control code or learned policy.
Some name the purpose of a move, such as \code{approach}, \code{transport} or \code{retreat}.
Some merge several moves into one call, such as \code{pick}, which moves, closes and lifts, or \code{handover}, in which one arm presents an object and the other takes it.
And \code{adjust\_view} moves a wrist camera to see better, with guidance to look before acting.
The model used the named operations in most of its calls, and they changed what it did more than how well it did it.
It moved its cameras more than three times as often and dropped more of what it held, and the score did not rise.

\paragraph{Move sequences.}
Allowing up to four moves per decision raised the score, mostly on Long-horizon tasks.
The gain has a simple source.
Episodes need fewer decisions, and the few episodes that used to run out of the budget of 100 decisions account for most of the difference (\cref{tab:moveseq} in \cref{app:interfaces}).
The gain therefore reflects the budget more than the interface, and the final configuration of \justrua removes the budget instead.

\subsection{Reasoning Effort}
\label{sec:effort}

Medium effort, the default, scores highest overall.
Low effort costs slightly less and gives up most of the Precision score.
High effort costs more and trades precision for memory.
The two other settings play differently enough from medium that we used low effort alongside medium when building the evaluation set (\cref{sec:evalset}), so that the choice of tasks does not rest on one style of play.

\takeaway{1}{A lean harness goes a long way.}{%
Keeping the full episode history and stopping each motion after 20 control steps lift the score from 29.7 to 38.0.
Interfaces that structure the action for the model, such as relative targets, named skills and model-chosen step counts, add nothing it cannot handle with plain absolute targets.}

\section{Reusing Experience}
\label{sec:experience}

Experience from other episodes is the core of recursive self-improvement (RSI) \citep{zhou2026whoimproves}, and of everything we tried it carries the greatest risk of gaming the benchmark.
It draws on the tasks being scored, so its scores are not comparable with a standard evaluation.
This is not to say that iteration is useless.
A score obtained with experience from the tasks being scored simply reports something else, and the setting as a whole is not one that a regular evaluation is meant for.
We test experience reuse because it is widely used, in the three forms of \cref{sec:axes}, and we report what each changes in behaviour as well as in score.
In our runs no form clearly beat the default (\cref{tab:main}), and none of them comes for free.

\subsection{Demonstrations}
\label{sec:demos}

Demonstrations placed in context are a common way to help GPT-6 Astra \citep{cai2026sensenova,liu2026roboicl}.
Both of our forms draw on one expert-curated successful episode per task from RoboDawn \citep{guo2026robodawn}, looked up by task name and inserted once before the first decision.
The \emph{visual demonstration} shows the 16 keyframes its author selected, with no action trace or robot state attached.
The \emph{text trajectory} gives the same episode stage by stage, as the poses of both arms, the commands issued and their outcomes, converted to our coordinates, with failed commands kept and the author's plans left out.
It takes about 16,000 characters per task on average (\cref{app:demos}).
In both forms the model is told that the demonstration comes from a different episode and that it should follow its order and geometry rather than copy its coordinates.

\begin{table}[!htb]
\centering
\caption{\textbf{Behaviour with reused experience}, counted over the same 42 first-round episodes.
The visual demonstration and the text trajectory are looked up by task name, and the self-built library is not.}
\label{tab:experience}
\small
\setlength{\tabcolsep}{7pt}
\begin{tabular}{@{}lcccc@{}}
\toprule
& Default & \makecell{Visual\\demonstration} & \makecell{Text\\trajectory} & \makecell{Self-built\\library} \\
\midrule
Handovers through the table & 9 & 15 & 13 & 9 \\
Dropped objects & 26 & 20 & 17 & 22 \\
Arm-to-arm collisions & 6 & 0 & 0 & 0 \\
Empty grasps & 47 & 50 & 53 & 33 \\
\midrule
Decisions per episode & 67.7 & 65.8 & 71.1 & 60.9 \\
Cost per episode (\$) & 4.94 & 5.11 & 5.84 & 5.43 \\
\bottomrule
\end{tabular}
\end{table}

With either demonstration the model passed objects through the table more often and dropped fewer of them (\cref{tab:experience}).
On the evaluation set neither moves the overall score clearly in either direction (\cref{tab:main}).

\begin{figure}[!htb]
\centering
\includegraphics[width=0.94\linewidth]{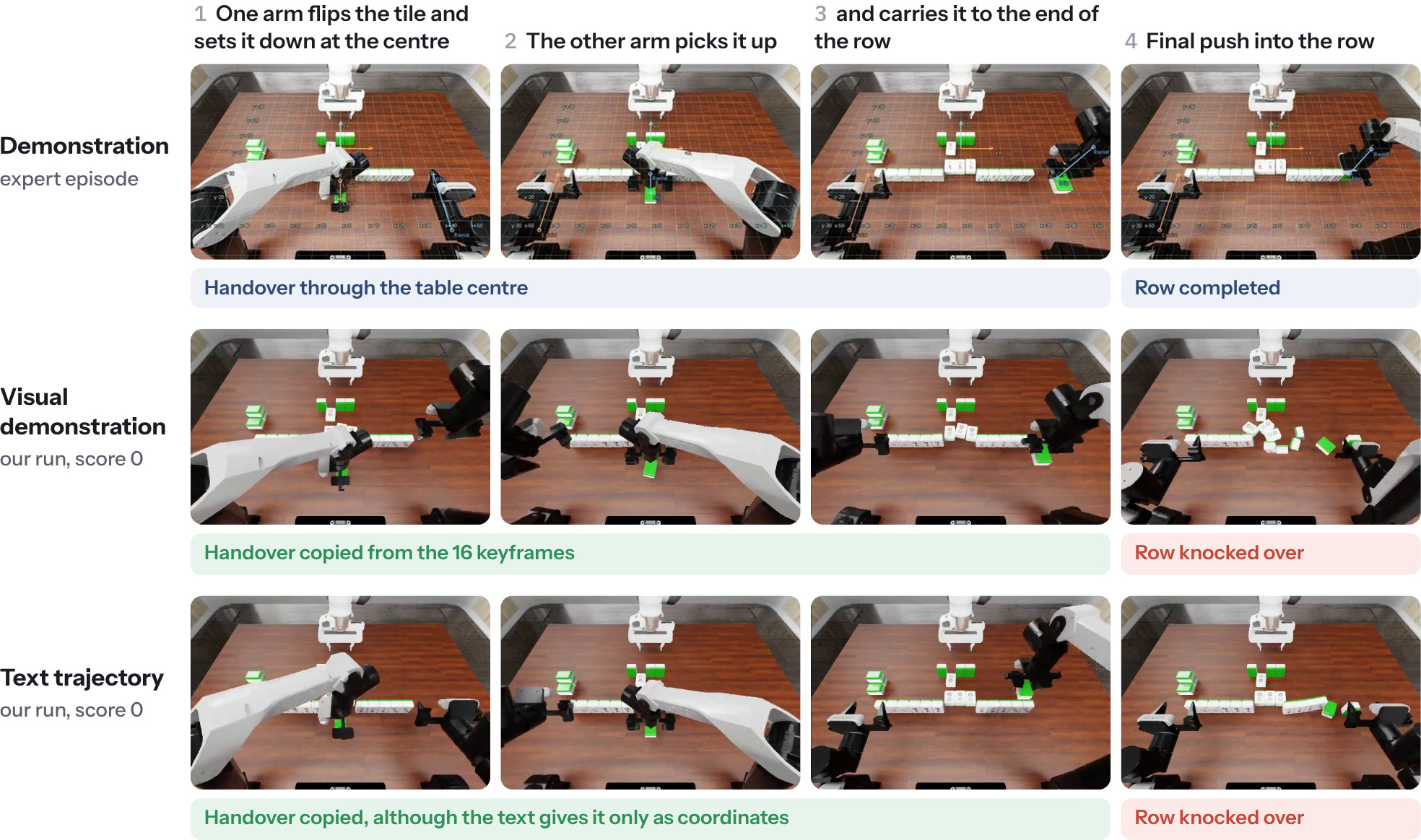}
\caption{\textbf{\code{make\_kong} with a demonstration}, head camera.
Top, four keyframes of the expert episode.
Middle and bottom, the same moments in our runs with the visual demonstration and with the text trajectory.
Both runs copy the handover through the table centre.
Neither copies the last step, where the demonstration grips the tile upright to push it into the row, and both knock the row over and score 0.}
\label{fig:kong}
\end{figure}

\paragraph{The \code{make\_kong} case.}
\code{make\_kong} shows why a better plan is not enough.
In this task a tile has to be flipped and added to a row, and a single arm cannot reach the flip angle the tile needs and still place it.
Without a demonstration the model never hands the tile to the other arm (\cref{sec:errors}).
With one it copies the handover (\cref{fig:kong}).
One arm flips the tile and sets it down at the table centre, and the other carries it to the end of the row.
The text trajectory carries this handover only as coordinates, the left hand opening at a point where the right hand later closes, and the model still found it.
Neither run copies the last step, where the demonstration grips the tile upright to push it into the row, and both lose the tiles on that push and score 0.
The demonstration fixed the plan, and the episode then failed somewhere else.
Fixes of this kind are lost to failures elsewhere, so the overall score barely moves while each episode costs more.

\subsection{Self-Improvement Loop}
\label{sec:loop}

Several RoboDojo entries improve themselves by writing and refining their own skills \citep{li2026embodiedevo,sensenova2026roborsi}.
In our loop a Codex agent writes extra operations for the model, as Python code built on \code{move\_eef}, together with a guide on when to use them.
It starts from the RoboDawn demonstrations, from which it writes a first library of four operations, \code{grasp}, \code{transport}, \code{place} and \code{push}.
The tasks are split into five groups.
In each round the model runs the current library on one group, and the agent then reviews those runs, with their scores, action logs and videos, and revises the library for the next round.
Nothing is edited by hand.
Unlike the demonstrations, the library is not stored or looked up by task, and every task sees the same operations and the same guide.

The agent's changes follow the failures it saw (\cref{tab:loop} in \cref{app:loop}).
After the model let go of a plug and a pen before they were in place, \code{place} and \code{grasp} gained an option to stop before opening or closing, and a new \code{insert} pushes a held tip into a hole once it is lined up.
After the model counted presses that never registered in \code{play\_Xylophone} and \code{swap\_blocks}, the guide began to separate attempted presses from confirmed ones.
Most changes make an operation stop earlier so that the model can check the result before going on, which echoes the 20-step cap of \cref{sec:motion}.

\begin{table}[!htb]
\centering
\caption{\textbf{The self-built library after five rounds.}
Each task is practised once inside the loop.
The iterate set holds the tasks whose runs the loop recorded, scored once inside the loop and once with the final library.}
\label{tab:loop-summary}
\small
\setlength{\tabcolsep}{7pt}
\begin{tabular}{@{}lccl@{}}
\toprule
& Reference & Final library & Reference is \\
\midrule
Evaluation set & 38.0 & 37.0 & the default testbed \\
Iterate set & 35.4 & 36.4 & the runs inside the loop \\
Empty grasps, first round & 47 & 33 & the default testbed \\
Arm-to-arm collisions, first round & 6 & 0 & the default testbed \\
\bottomrule
\end{tabular}
\end{table}

The library changed how the model grasped, with fewer empty grasps and no arm-to-arm collisions in the first round (\cref{tab:experience,tab:loop-summary}).
With each task practised once, it scored slightly higher than the loop's own runs on the tasks it had seen and stayed level with the default on the evaluation set.
The loop helps, but slowly.

\subsection{Cost of Experience}
\label{sec:experience-discussion}

The loop ran for five rounds and practised each task once, so we cannot rule out larger gains at a larger scale.
Systems that report clear gains from self-improvement practise each task many more times than we did \citep{li2026embodiedevo,li2026physrsi}.
Within that limit, the three forms share one pattern.
Experience corrects the behaviour it targets.
Demonstrations made handovers more frequent and drops rarer, and the library made grasps land more often.
But a task draws on many abilities at once, and correcting one of them rarely changes the outcome, as \code{make\_kong} shows.

The gains also come at a price.
Every episode carries the extra context, which raises its cost (\cref{tab:experience}), and the cost of producing the experience, expert episodes for the demonstrations and practice runs for the library, comes on top.
The experience has to come from the tasks being scored, which is what makes its scores hard to compare with a standard evaluation.
For these reasons \justrua keeps no form of experience reuse, whatever its score.

\takeaway{2}{Reusing experience is not a free lunch.}{%
Demonstrations and a self-written skill library change how the model acts, but neither clearly beats the default, and both make an episode more expensive.
They also draw on the tasks being scored, so their scores are not comparable with a standard evaluation.}

\section{Scene Geometry}
\label{sec:geometry}

Recent agent systems feed the model depth, point clouds and affordances from dedicated models \citep{qu2026roboharness,zhou2026gptgear,huang2026rax,kim2026claro,yang2026ora,wang2026auriga}.
We add scene geometry in four steps and ask how much of it GPT-6 Astra actually needs.
Of the four axes this one helps most (\cref{tab:main}), and the behaviour counts say why (\cref{tab:geometry}).
The model's own grip readings, depth predicted from RGB and a little code that turns pixels into 3D coordinates go a long way without simulator ground truth, while precomputed affordances, the most elaborate piece, did not help.

\begin{table}[!htb]
\centering
\caption{\textbf{Behaviour and added context with scene geometry}, over the same 42 first-round episodes.
Each block adds to the kept row before it.
Added context is in characters per decision for robot readings and in tokens per observation otherwise.
Episodes of the final configuration have no decision cap, so its first row counts events within the first 100 decisions, as for the rows above, and its second row whole episodes.
Shaded rows are kept in \justrua.}
\label{tab:geometry}
\small
\setlength{\tabcolsep}{5pt}
\begin{tabular}{l cccc c}
\toprule
Configuration & \makecell{Empty\\grasps} & \makecell{Fine placement\\failures} & \makecell{Hand-\\overs} & \makecell{Free-arm\\views} & \makecell{Added\\context} \\
\midrule
\grouprow{6}{Robot readings}
None (default testbed) & 47 & 20 & 15 & 34 & 0 \\
Full JSON & 47 & 23 & 11 & 32 & 3,100 \\
\kept Compact text & 41 & 18 & 13 & 32 & 1,800 \\
\midrule
\grouprow{6}{Depth}
Simulator depth (reference) & 17 & 15 & 9 & 25 & 3,500 \\
\kept Predicted depth & 16 & 7 & 15 & 31 & 3,700 \\
\midrule
\grouprow{6}{Affordances}
Drawn, simulator depth & 17 & 15 & 14 & 29 & 3,000 \\
Drawn with RGB and depth, simulator & 25 & 27 & 11 & 27 & 5,300 \\
Drawn with depth, predicted & 39 & 14 & 9 & 39 & 3,200 \\
Affordance panel, predicted depth & 34 & 16 & 14 & 40 & 3,300 \\
\midrule
\grouprow{6}{Code}
\kept Coordinate solving (\justrua) & 15 & 7 & 11 & 28 & \na \\
\quad whole episodes & 20 & 10 & 14 & 30 & \na \\
\bottomrule
\end{tabular}
\end{table}

\subsection{Robot Readings}
\label{sec:readings}

Robot readings are the basic sensor values of each gripper, finger positions and forces, contact force and wrist load.
They carry no privileged information, such as the true pose of any object.
Their clearest use is in grasping.
The default interface reports the jaw opening the model asked for, so after closing on an object the model sees its own target.
The readings give the opening the fingers actually reached and the force at each fingertip, which tells the model whether the object is held (\cref{tab:readings}).

\begin{table}[!htb]
\centering
\begin{minipage}[t]{0.45\linewidth}
\centering
\caption{\textbf{Robot readings after a grasp} in \code{insert\_tubes}.
The default shows the commanded opening of 0.24.
The fingers stopped at 0.44 on the same scale and press on both sides, so the tube is held.}
\label{tab:readings}
\footnotesize
\setlength{\tabcolsep}{3.5pt}
\begin{tabular}{@{}lcc@{}}
\toprule
After closing & Default & With readings \\
\midrule
Jaw opening, 0 to 1 & 0.24 & 0.44 (13.6\,mm) \\
Contact force (N) & \na & 24.9 / 24.5 \\
\bottomrule
\end{tabular}
\end{minipage}\hfill
\begin{minipage}[t]{0.51\linewidth}
\centering
\caption{\textbf{Error of predicted depth} against simulator depth in the table area, averaged over six head-camera frames from \code{make\_kong} and \code{fill\_egg\_holder}.
$r$ and $\rho$ are the Pearson and Spearman correlations, which scaling a frame leaves unchanged.}
\label{tab:depth-error}
\footnotesize
\setlength{\tabcolsep}{3.5pt}
\begin{tabular}{@{}lccc@{}}
\toprule
Head camera & MAE & $r$ & $\rho$ \\
\midrule
Raw prediction & 73.7\,cm & 0.993 & 0.992 \\
True focal length & 40.1\,cm & 0.993 & 0.992 \\
Scaled per frame & 7.3\,cm & 0.993 & 0.992 \\
\bottomrule
\end{tabular}
\end{minipage}
\end{table}

How the readings are written matters as much as what they contain.
Information about the scene arrives at every decision and stays in the history, so every extra character is paid many times over.
Written as full JSON, with vectors, thresholds and identifiers for every finger, the readings crowded the context and pushed many first-round episodes into context compaction, which the default testbed never reaches.
\justrua writes them in a short text format with selected fields, fixed units and rounded values (\cref{lst:readings} in \cref{app:geometry}).
In this form no first-round episode was compacted, and empty grasps fell slightly (\cref{tab:geometry}).

\subsection{Depth}
\label{sec:depth}

Many recent agent systems use depth to place objects in 3D \citep{qu2026roboharness,zhou2026gptgear,huang2026rax}, but the RoboDojo leaderboard \citep{robodojo2026leaderboard} does not accept depth read from the simulator for agent entries, as a maintainer stated in review \citep{xpolicylab2026review}.
We therefore test both, depth from the simulator and depth that Depth Pro \citep{bochkovskii2025depthpro} predicts from each RGB image alone.
In both cases each RGB image is paired with a depth image that shows camera-forward distance on a fixed colour scale, and the text adds the pose and intrinsics of each camera and a few more gripper readings, such as the finger gap and the contact points.
Depth Pro receives nothing but the RGB image.
It estimates its own focal length, and we pass it no camera parameters and fit no scale to its output.
The instruction tells the model that the predicted depth is an estimate and is least reliable around thin objects and occlusion boundaries.

\begin{figure}[!htb]
\centering
\includegraphics[width=\linewidth]{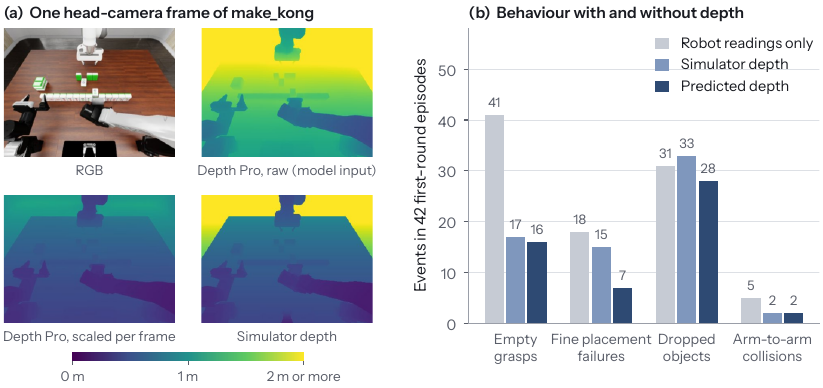}
\caption{\textbf{Predicted depth.}
(a) One head-camera frame with the raw Depth Pro prediction that the model receives, the same prediction scaled to the robot geometry and simulator depth, on a shared colour scale.
(b) Behaviour counts with robot readings only, with simulator depth and with predicted depth.}
\label{fig:depth}
\end{figure}

The predicted distances are far off, by tens of centimetres even with the true focal length (\cref{tab:depth-error}).
A scale fitted to the robot's own geometry in each frame brings the error down to a few centimetres, but that scale drifts between frames and cannot be fitted for the wrist cameras, so \justrua passes the raw predictions on (\cref{fig:depth}a).

Even so, predicted depth improves manipulation almost as much as simulator depth.
Empty grasps fell sharply with either source, failed fine placements fell further with predicted depth, and drops did not fall with either (\cref{fig:depth}b).
What the prediction gets right is the order of depths, and that seems to be enough.
The gain from depth does not need privileged information.

\subsection{Affordances}
\label{sec:affordances}

Depth tells the model where surfaces are, but not where an object can be held or which arm can reach it.
That knowledge is an affordance.
Recent agent systems compute affordances with dedicated models and hand them to the planner \citep{kim2026claro,yang2026ora,li2026roboharnessk1}, and since many episodes go wrong at the grasp, we tested whether giving the model this knowledge directly would help.
A pretrained grasp generator, GraspGen \citep{murali2025graspgen}, proposes up to eight grasp poses on the point cloud of each object, and inverse kinematics tests which surfaces each arm can reach with its current tool orientation.
Both are drawn onto the camera images, and the text lists the grasp poses along with hints such as which candidates a camera cannot see.

\begin{figure}[!htb]
\centering
\includegraphics[width=\linewidth]{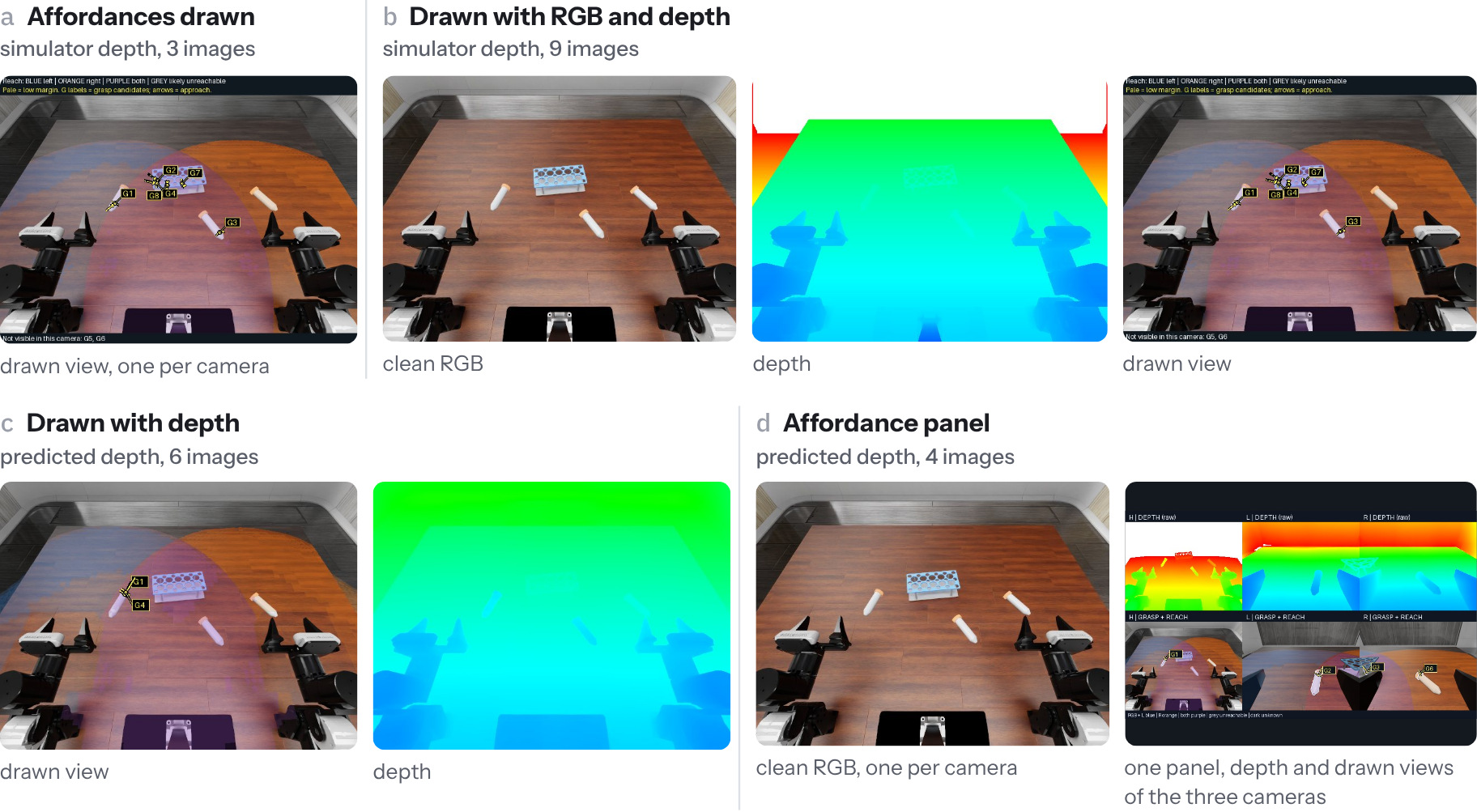}
\caption{\textbf{Four ways to present affordances}, head-camera images of one observation of \code{insert\_tubes} as the model receives them.
Yellow labels are grasp candidates with their approach direction, and blue, orange and purple shading marks surfaces the left arm, the right arm or both arms can reach.
(a) Marks drawn onto each camera view, with no depth image.
(b) Clean RGB and depth added back, so each camera appears three times.
(c) Predicted depth, with the clean RGB left out again.
(d) Clean RGB at full size and one panel that holds the six other views at $256\times192$ pixels, with at most four grasp candidates.}
\label{fig:affordances}
\end{figure}

The real difficulty lies in presenting all this.
Each observation already carries the camera views and their depth, and anything added either takes more space or crowds what is already there.
We tried four presentations (\cref{fig:affordances}), which differ in whether the marks are drawn over the camera view or shown next to it, and in how many images an observation then holds.

Against depth images from the same source, none of the four grasped more reliably or scored higher (\cref{tab:main,tab:geometry}).
The best of them, the affordance panel, matches predicted depth in score with about twice as many empty grasps.
The moments where affordances could make a difference are also few.
Reach matters most when one arm cannot get to an object and hands it to the other, and handovers stayed rare in every version.
Grasp candidates matter only in the decisions that start a grasp.
The cost, in contrast, comes with every observation.
Information that is needed only now and then is a poor fit for an observation that is paid for at every decision, and we left affordances out of \justrua.

\subsection{Coordinate Solving}
\label{sec:coord}

In the last step the model may write and run code in its Codex session, as in Code as Policies \citep{liang2023cap} and, more recently, URAI \citep{ge2026urai} and HexaShell \citep{gao2026hexashell}.
Web search stays off so that the model cannot look up the benchmark or its solutions, and the simulator waits while the code runs.
At the same time the 100-decision cap that all earlier comparisons keep is lifted, while the official limit of 500 control steps still applies.
The step therefore changes two things together, and we report it as one.

\begin{figure}[!htb]
\centering
\includegraphics[width=\linewidth]{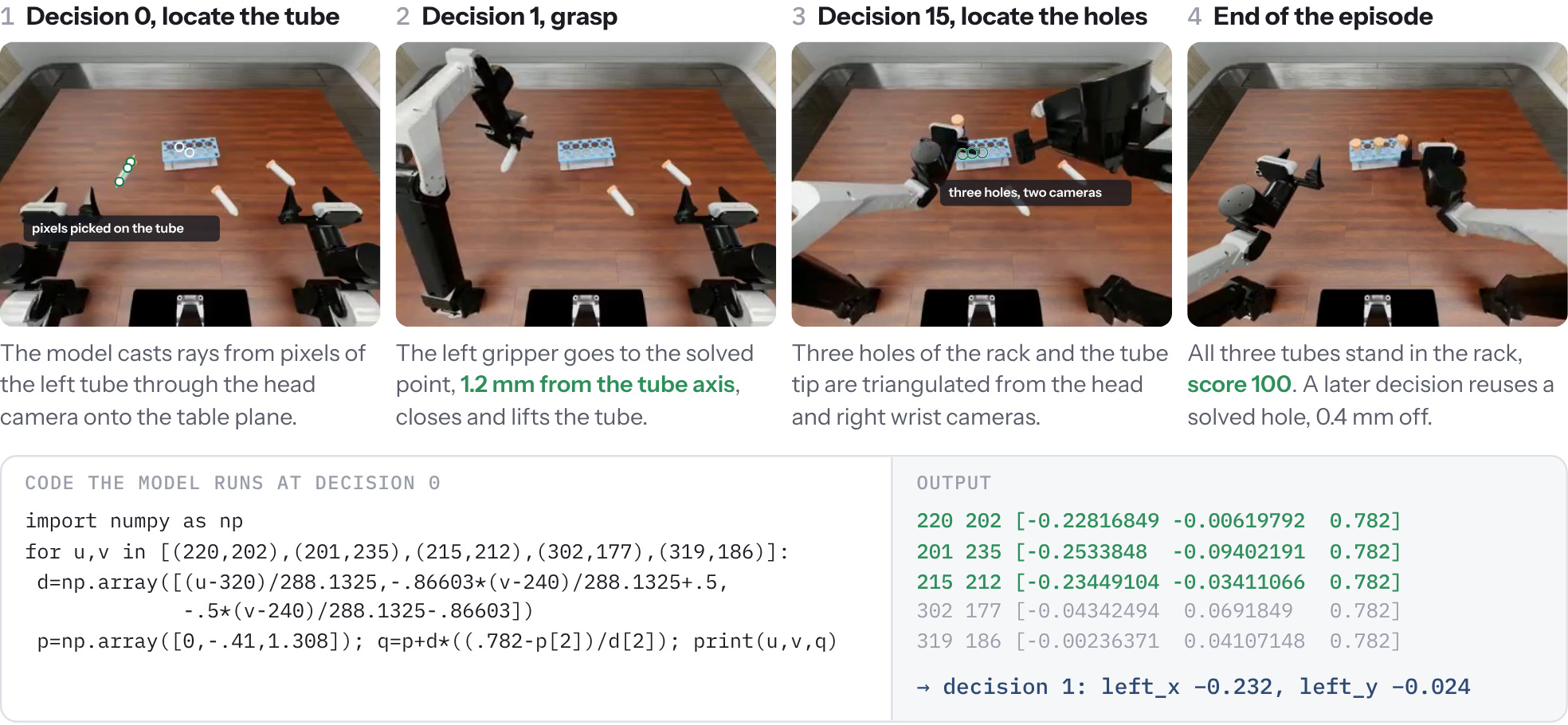}
\caption{\textbf{Coordinate solving in \code{insert\_tubes}}, head camera.
At its first decision the model casts rays from pixels of the left tube through the head camera onto the table plane, and its next command lands 1.2\,mm from the solved tube axis.
At decision 15 it triangulates three holes of the rack and the tube tip from two cameras.
Bottom, the code of the first run and its output.
Distances compare the output of the code with the commands that followed, not with the true positions of the objects.
Two more cases are in \cref{fig:solver-more}.}
\label{fig:coord}
\end{figure}

The model used code in a consistent way.
Almost every code run in the first round solves for coordinates (\cref{tab:code-use}).
The model casts rays from pixels through the camera calibration and then triangulates a point seen by two cameras or meets a known surface height.
The code works from the images and text the model already receives and never reads the depth images.
Code therefore acts as one more source of scene geometry, which is why we place it on this axis.

\begin{table}[!htb]
\centering
\begin{minipage}[t]{0.55\linewidth}
\centering
\caption{\textbf{Code runs} in the 42 first-round episodes of \justrua, reviewed by hand.}
\label{tab:code-use}
\small
\begin{tabular}{@{}lcc@{}}
\toprule
Use of code & Runs & Share \\
\midrule
Geometry, e.g.\ triangulation & 123 & 93.2\% \\
Listing inputs or rereading its own code & 5 & 3.8\% \\
Cropping and enlarging an image & 3 & 2.3\% \\
No output & 1 & 0.8\% \\
\midrule
Total & 132 & 100\% \\
\bottomrule
\end{tabular}
\end{minipage}\hfill
\begin{minipage}[t]{0.41\linewidth}
\centering
\caption{\textbf{One layout of \code{insert\_tubes}}, predicted depth against coordinate solving.}
\label{tab:tubes}
\small
\begin{tabular}{@{}lcc@{}}
\toprule
Same layout & Depth & Code \\
\midrule
Score & 40 & 100 \\
Decisions & 65 & 57 \\
Control steps, of 500 & 500 & 496 \\
\bottomrule
\end{tabular}
\end{minipage}
\end{table}

In \code{insert\_tubes}, for example, the model locates the tubes and the holes of the rack from pixels and the camera calibration (\cref{fig:coord}) and places all three tubes, on a layout where predicted depth alone left the task unfinished (\cref{tab:tubes}).
It did so in fewer decisions and within the same limit of control steps, so the gain on this layout comes from better targets and not from a longer episode.
Across the evaluation set the gain is largest on Long-horizon and Memory tasks, where episodes need many decisions (\cref{tab:main}).
It is also the most expensive step, since episodes run longer and every code run adds to the history.
Within the first 100 decisions, grasping and placing stay at the level predicted depth had reached (\cref{tab:geometry}).

\takeaway{3}{Scene geometry helps most.}{%
Depth images cut empty grasps from 41 to 16, even when Depth Pro predicts the depth from RGB alone and is off by tens of centimetres.
With code that solves 3D coordinates from pixels and no decision cap, the score reaches 50.3.
Precomputed affordances, the most elaborate form of geometry we tried, did not help.}

\section{Error Analysis}
\label{sec:errors}

\begin{figure}[!htb]
\centering
\includegraphics[width=0.96\linewidth]{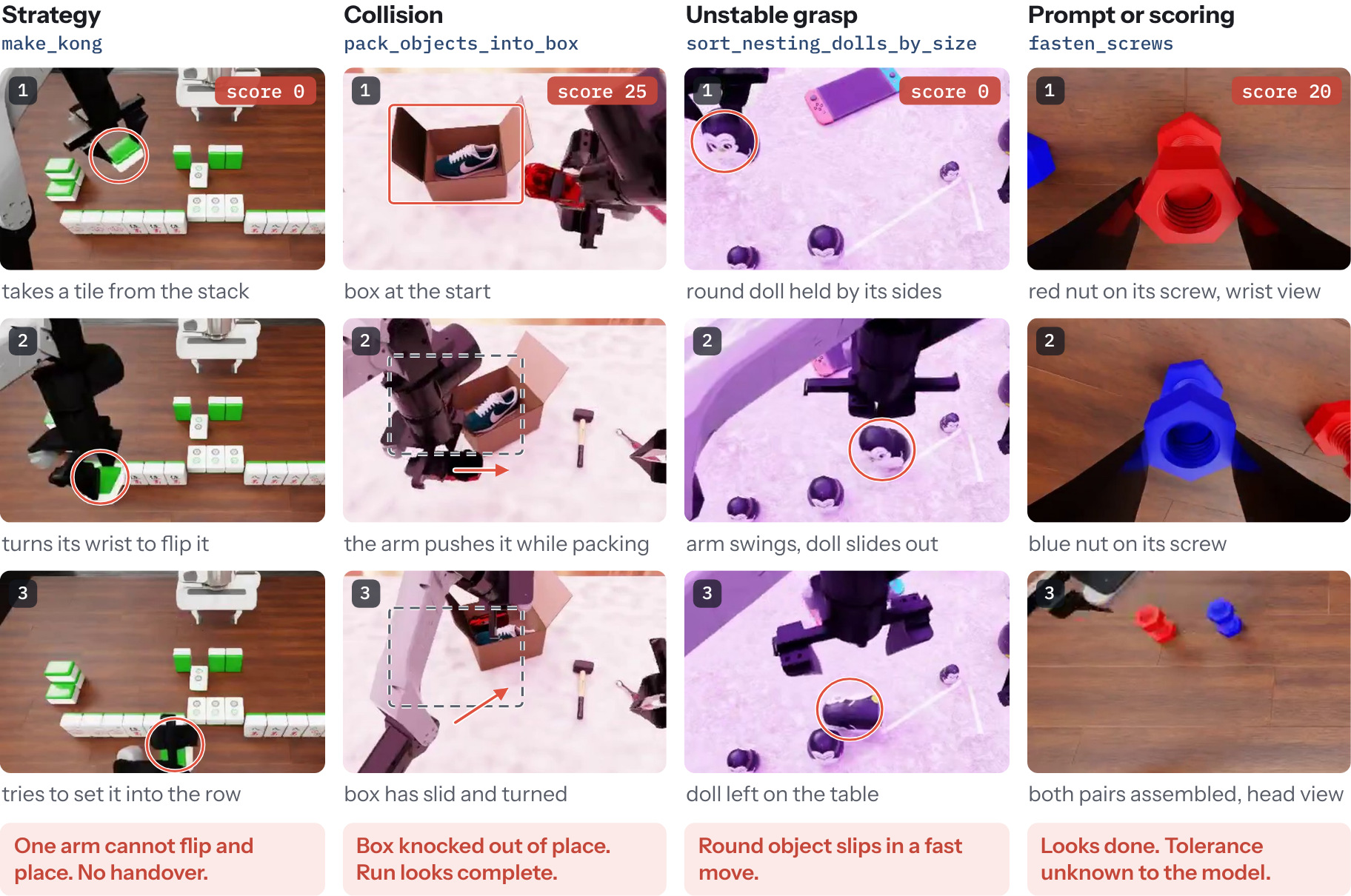}
\caption{\textbf{One failure of each cause}, three moments of the episode from top to bottom, with its first-round score.
Frames are from the head camera except where noted.
In \code{make\_kong} one arm tries to flip the tile and set it into the row by itself.
In \code{pack\_objects\_into\_box} the dashed outline marks where the box stood before the arm pushed it.
In \code{sort\_nesting\_dolls\_by\_size} the doll slides out of the gripper as the arm swings.
In \code{fasten\_screws} both pairs end up assembled, and the score stays low.}
\label{fig:error-cases}
\end{figure}

With the final configuration, 21 of the 42 first-round tasks score below 50.
Whether these failures share a cause decides what further gains would take, one more general fix or work on each task.
We watched every such episode and sorted the failures by cause (\cref{tab:errors,fig:error-cases}).
Some episodes fail for more than one reason, and most failures turn out to be specific to one task (\cref{tab:error-tasks} in \cref{app:errors}).

\begin{table}[!htb]
\centering
\caption{\textbf{Causes of failure} in the 21 first-round tasks where \justrua scores below 50, with the number of tasks for each cause and a direction that could address it.
A task can have more than one cause.}
\label{tab:errors}
\small
\renewcommand{\arraystretch}{1.12}
\begin{tabularx}{\linewidth}{@{}l >{\raggedright\arraybackslash}X c >{\raggedright\arraybackslash}X@{}}
\toprule
Cause & Definition and example & Tasks & Potential fix \\
\midrule
Strategy & The plan does not suit the arm, e.g.\ one arm cannot flip a tile and place it at once (\code{make\_kong}). & 8 & Planning with high-level knowledge of the arm and its limits. \\
Collision & Unplanned knocks, as the model misjudges the size of the arm or of what it holds (\code{pack\_objects\_into\_box}). & 6 & Collision checking with complete assets of the arm and held objects. \\
Unstable grasp & Irregular or round objects slip out of the gripper, such as a nesting doll. & 6 & Grasp planning on the same assets. \\
Prompt or scoring & The task looks easy, but the prompt or the scoring seems to be at fault (\code{fasten\_screws}). & 5 & Revising the task prompts, which raises the first-round score from 54.0 to 55.6. \\
\bottomrule
\end{tabularx}
\end{table}

\paragraph{Four causes.}
Strategy failures come from plans that do not suit the arm.
In \code{make\_kong}, a single arm cannot reach the flip angle the tile needs and still place it correctly.
Nothing tells the model this about the arm, so it never splits the job with a handover.
The pouring tasks fail in a similar way, since the model pours forward, a motion the arm does poorly, where a side pour would work.
Collisions are unplanned knocks, such as the arm pushing a light box out of place, after which the run looks complete but scores low.
Unstable grasps occur with irregular or round objects, such as a doll that slips when the arm moves too fast, or a coin that offers few stable grasp points.
The remaining cases look easy, and the prompt or the scoring seems to be at fault.
In \code{fasten\_screws}, for instance, the screws go in, but the model may be unable to tell whether it met the tolerance the scorer requires.

\paragraph{Possible fixes.}
Each of the first three causes points to a general direction, but each needs knowledge that is costly to supply.
Planning would need high-level knowledge of the arm, such as which flip angles one arm cannot reach while placing, and which limits matter changes from task to task.
Collision checking would need complete digital assets of the arm and of every object it holds, and modelling every held object at that level is hardly feasible.
Grasp planning would need the same assets to model how each irregular object sits in the gripper.
For the prompt and scoring cases, rewriting the recipes of the five tasks so that they state the conditions the scorer checks raises the first-round score slightly (\cref{tab:errors}, \cref{app:errors}).
The gain is small because these tasks are not all easy in other respects, and \code{pack\_objects\_into\_box}, for one, also fails by collision.
This revision draws on the scorer of each test task, so we report it here and do not count it in \justrua.

The other failures are hard to remove without trial and error on each task, which comes close to tuning on the test tasks themselves.
We read this as a sign that, for a general harness with no task-specific component, the remaining headroom on this benchmark lies mostly in its long tail.

\takeaway{4}{Remaining failures are mostly long-tail.}{%
Most failures left in our best configuration are specific to one task, such as plans that do not suit the arm, unplanned knocks and slipping grasps.
Fixing them would take trial and error on each task, which comes close to tuning on the test tasks.}

\section{Related Work}
\label{sec:related}

\paragraph{Frontier models as robot policies.}
Vision-language-action models learn robot control from large robot datasets on top of pretrained vision-language backbones \citep{brohan2023rt2,kim2024openvla,black2024pi0,pi2025pi05,pi2026pi07,yu2026dm0}.
A more recent line drives the robot with a general-purpose model or agent that receives no robot training at all \citep{jia2026agp}.
For GPT-6 Astra, RoboProbe \citep{zhang2026roboprobe} provides the first evaluation on RoboDojo together with the official L3 harness, Galbot \citep{galbot2026exploring} and SenseNova \citep{cai2026sensenova} study its abilities more broadly, and Astranomy \citep{yang2026astranomy} treats it as a native robot policy with action chunking.
Systems such as RoboDawn \citep{guo2026robodawn}, GPT-GEAR \citep{zhou2026gptgear}, RoboHarness \citep{qu2026roboharness}, ORA \citep{yang2026ora} and uni-robot-agent \citep{yuan2026unirobotagent} each build a harness with its own interfaces and perception.
We do not propose another system.
We test which of their components matter when all of them are measured under one protocol, and the configuration we end with is a by-product of that test.

\paragraph{Skills and code.}
Named skills are a common way to connect language models to robots \citep{ahn2022saycan}, and code lets a model compose perception and control by itself \citep{liang2023cap,huang2023voxposer,fu2026capx}.
URAI \citep{ge2026urai} lets frontier agents write, call and evolve their own robot tools, and CLARO \citep{kim2026claro} orchestrates open-vocabulary grasping with VLM agents and dedicated models.
Our semantic operations and coordinate solving follow these two lines while using only what the official harness already provides.
We find the two to be of unequal value for GPT-6 Astra.
Skills that restate its moves at a higher level add little, while code that computes where to move adds a lot.

\paragraph{Experience and self-improvement.}
Language agents can improve across attempts by storing skills \citep{wang2023voyager} or verbal reflections \citep{shinn2023reflexion}, and ASPIRE \citep{lu2026aspire} builds such a skill library for robots.
On RoboDojo, several systems refine their own skills over many rounds \citep{li2026embodiedevo,sensenova2026roborsi,yu2026robodoctor} or scale online exploration \citep{huang2026rax}, and PhysRSI-eval \citep{li2026physrsi} studies where the gains of such loops come from.
These systems practise each task many times.
We test a lighter regime, a single pass over the tasks, and look at what experience changes in behaviour as well as in score.

\paragraph{Controlled studies.}
Our study follows a line of work that explains a result by taking a system apart rather than by adding to it, in the spirit of JustRL \citep{he2026justrl}.
For software agents, the design of the agent-computer interface is known to change what a model can do \citep{yang2024sweagent}, and a study of Claude models on robotics tasks reports that performance depends strongly on how the model is connected to the robot \citep{berman2026claudeplaysrobotics}.
Astranomy \citep{yang2026astranomy} does so for GPT-6 Astra as a native robot policy, and PhysRSI-eval \citep{li2026physrsi} for self-improvement loops.
We cover the harness, the action interface, experience and scene geometry under one evaluation set, one behaviour review and one cost measure.

\section{Conclusion}
\label{sec:conclusion}

Starting from the official L3 harness, \justrua keeps three kinds of change.
The loop keeps the full episode history and stops each motion early, the observation carries robot readings and predicted depth, and the model may solve coordinates in code.
Together they raise the evaluation-set score from 29.7 to 50.3 with no learned policy, no self-improvement loop and no task-specific tuning.
Alternative action interfaces, precomputed affordances and reused experience did not earn their place under our protocol, and most of the remaining failures are specific to single tasks.

Two observations run through these results.
First, GPT-6 Astra handles the plain interface well.
World-frame targets, raw sensor values and uncalibrated depth are enough for it, and layers that restructure its actions or pre-digest its observations mostly add cost.
Second, what helps is information that bears on the errors the model makes often, such as closing on nothing or misjudging where an object is, delivered in a form short enough to pay for at every decision.

\paragraph{Limitations.}
All experiments run in simulation, on one benchmark and with one model, and each configuration is run once on the evaluation set.
Each alternative is tested at the point of the path where it was added, not in every combination.
The experience runs draw on the tasks being scored, and the self-improvement loop ran for only five rounds with one practice run per task.
The evaluation set was chosen from the history of our own default testbed, and it weights random layouts more heavily than the official protocol does in the Generalization group, so its scores are not leaderboard scores.

\bibliographystyle{iclr2027_conference}
\bibliography{main}

\clearpage
\appendix

\section{Variants on Each Axis}
\label{app:axes}

\begin{table}[!ht]
\centering
\caption{\textbf{What we varied on each axis.}
Shaded rows are kept in \justrua, and the official L3 harness is the starting point.
\Cref{tab:main} gives the score of each variant.}
\label{tab:axes}
\small
\begin{tabularx}{\linewidth}{@{}l l >{\raggedright\arraybackslash}X@{}}
\toprule
Axis & Variant & Description \\
\midrule
\multirow{4}{*}{\makecell[tl]{Harness\\[1pt]\footnotesize\textit{the loop around the model}}}
& \cellcolor{keptrow}Official L3 & \cellcolor{keptrow}Reference loop, keeps the images of the last two observations \\
& \cellcolor{keptrow}Full history & \cellcolor{keptrow}Codex session that keeps every image of the episode \\
& Low effort & Reasoning effort below the default, medium \\
& High effort & Reasoning effort above the default \\
\midrule
\multirow{5}{*}{\makecell[tl]{Action\\[1pt]\footnotesize\textit{how a decision}\\\footnotesize\textit{becomes motion}}}
& \cellcolor{keptrow}20-step cap & \cellcolor{keptrow}Each motion stops after at most 20 control steps \\
& Model-chosen steps & 3 to 32 steps per motion, set by the model \\
& Relative targets & Offsets from the measured pose \\
& Semantic operations & \code{approach}, \code{grasp}, \code{place} and others, built on \code{move\_eef} \\
& Move sequences & Up to four moves per decision \\
\midrule
\multirow{3}{*}{\makecell[tl]{Reusing experience\\[1pt]\footnotesize\textit{what carries over}\\\footnotesize\textit{from other episodes}}}
& Visual demonstration & 16 keyframes of one expert episode, looked up by task \\
& Text trajectory & The same episode as poses, commands and outcomes \\
& Self-built library & Operations the system writes over five rounds \\
\midrule
\multirow{5}{*}{\makecell[tl]{Scene geometry\\[1pt]\footnotesize\textit{what the model is told}\\\footnotesize\textit{about the 3D scene}}}
& \cellcolor{keptrow}Compact readings & \cellcolor{keptrow}Finger positions and forces, contact force and wrist load, rounded \\
& \cellcolor{keptrow}Predicted depth & \cellcolor{keptrow}One depth image per camera, predicted from RGB \\
& Simulator depth & The same, read from the simulator, as a reference only \\
& Affordances & Grasp candidates and arm reach, in four presentations \\
& \cellcolor{keptrow}Coordinate solving & \cellcolor{keptrow}Code execution for target coordinates \\
\bottomrule
\end{tabularx}
\end{table}

\section{Testbed Details}
\label{app:testbed}

\paragraph{Instruction and task recipe.}
Each episode opens with a fixed instruction that defines \code{move\_eef} and \code{give\_up} and the JSON format of a decision, followed by the task instruction and a task recipe.
The recipes come from the RoboDojo wiki through the official harness and give the capability group, a short description and the official scoring ladder, without an action sequence (\cref{lst:recipe}).
They are the only task-specific text in the system, and we use them unchanged except in the prompt-revision experiment of \cref{app:errors}.

\begin{lstlisting}[float=ht,caption={\textbf{Task instruction and recipe} for \code{play\_Xylophone}, as the model receives them.},label={lst:recipe}]
Goal: Pick up the mallet and strike all xylophone keys from left to right.

TASK RECIPE:
# Play Xylophone
## Capability dimension
Precision - Fine-grained manipulation tasks with tight spatial constraints.
## Description
There is a xylophone and a mallet. The robot needs to pick up the mallet with
one hand and strike all the xylophone keys from left to right.
## Scoring
| Score | Condition |
| 0   | The full left-to-right strike sequence is not completed. |
| 100 | The mallet tip hits all xylophone keys from left to right, and after
        each hit the mallet is lifted by at least 0.025 m. |
\end{lstlisting}

\paragraph{Observation and feedback.}
At each decision the model receives the three RGB views, the measured grasp-point pose, gripper opening and joint state of both arms, and feedback on the previous command, including how far the grasp point ended up from its target and, under the 20-step cap, how many planned steps ran.
At every step we also log the requested target, the measured state after execution and the remaining error, so that a failed grasp can be traced to a wrong target, a poor plan or an imprecise execution.

\paragraph{Budgets.}
An episode has three separate limits, the number of model decisions, the control steps executed per motion and the benchmark's own limit of 500 control steps.
All controlled comparisons keep 100 decisions per episode and, from the default testbed on, at most 20 control steps per motion.
The decision limit is lower than the 170 calls allowed in the published run of the official harness \citep{zhang2026roboprobe}.
The final configuration of \justrua lifts the decision limit and keeps the other two, and an episode ends when either remaining limit runs out.

\paragraph{The 20-step cap.}
The cap is a thin wrapper around the unchanged planner, and it follows a few fixed rules.
The original planner and its retiming run as before.
Only a planned path longer than 20 control steps is cut, and the first 20 steps run with their original values, so the trajectory is never resampled.
The rest of the path is discarded for good, including any gripper command in it, and the gripper returns to the last command that actually ran.
The feedback keeps the original target, reports the remaining error to it and states the exact number of steps that ran.
No collision filter, refusal rule, replay of the discarded part or alternative controller is added.

\paragraph{Episode history.}
The Codex session keeps every observation, action and feedback message of the episode, images included, until Codex compacts the context near its window limit.
In the default testbed this did not happen, since requests stayed well below the size at which Codex compacts.
Configurations that add long text to every observation can reach the limit, as the full-JSON readings of \cref{sec:readings} did in 38\% of first-round episodes.

\section{Action Interfaces}
\label{app:interfaces}

\Cref{tab:interfaces} defines the interfaces of \cref{sec:harness}, and \cref{lst:calls} shows one call of each kind at the same moment of \code{insert\_tubes}, approaching the left tube.

\begin{table}[!ht]
\centering
\caption{\textbf{Action interfaces.} All interfaces run on the same planner and simulator. The semantic operations add no control code or learned policy beyond \code{move\_eef}.}\label{tab:interfaces}
\footnotesize
\setlength{\tabcolsep}{3.5pt}
\begin{tabularx}{\linewidth}{@{}>{\raggedright\arraybackslash}p{0.14\linewidth} >{\raggedright\arraybackslash}p{0.25\linewidth} >{\raggedright\arraybackslash}X@{}}
\toprule
Interface & Tools & Semantics \\
\midrule
Absolute targets (default) &
\code{move\_eef(targets, note)}\newline \code{give\_up()} &
World-frame targets for the grasp point of either arm, \code{left\_x/y/z} in metres, \code{left\_pitch/roll/yaw\_deg}, and \code{left\_gripper} from 0 (closed) to 1 (open), with the same fields for the right arm. Both arms can move in one call. Unnamed dimensions keep their observed value. Orientation is measured from a fixed straight-down grasp rather than added to the current one, and a gripper command runs after the arm motion. \code{give\_up} ends an episode the model judges unrecoverable. \\
\midrule
Relative targets &
\code{move\_eef\_relative(}\newline \code{\ \ targets, note)}\newline \code{give\_up()} &
Increments \code{left\_dx/dy/dz} in metres and \code{left\_drotation\_x/y/z\_deg} from the full-precision pose measured at this decision, along and about fixed world axes. Missing increments are zero. Rotations apply about $x$, then $y$, then $z$, and keep the grasp point in place. The gripper stays absolute. Observations and recipes still report absolute positions. \\
\midrule
Semantic operations &
\code{approach}, \code{transport}, \code{retreat}, \code{grasp}, \code{pick}, \code{clamp}, \code{place}, \code{deposit}, \code{release}, \code{handover}, \code{pour\_step}, \code{adjust\_view}, \code{move\_eef}, \code{give\_up} &
Each call names the object it acts on, a label that records intent and does not steer motion, and expands into stages. Every stage runs through the original planner for at most 20 control steps, and the next stage starts only after the arm arrives within 1\,cm and 5$^\circ$. \code{approach}, \code{transport} and \code{retreat} move the named pose dimensions while the rest of the pose and the grip hold. \code{grasp} and \code{place} move and then close or open if asked. \code{pick} moves, closes and lifts, and \code{deposit} moves, opens and retreats, in one call. \code{clamp} and \code{release} set the jaw only. In \code{handover} the giver moves, the receiver opens, approaches, grasps and closes, and the giver keeps its grip until an explicit release. \code{pour\_step} is one pouring motion and \code{adjust\_view} one move of a wrist camera to see better. \\
\midrule
Model-chosen steps & \code{move\_eef(...,}\newline \code{\ \ num\_actions)} & As the default, with 3 to 32 control steps per motion chosen by the model in place of the fixed cap of 20. \\
\midrule
Move sequences & \code{move\_eef\_sequence(}\newline \code{\ \ num\_moves, moves)} & Up to four \code{move\_eef} moves per decision, executed in order. \\
\bottomrule
\end{tabularx}
\end{table}

\begin{lstlisting}[float=ht,caption={\textbf{One call of each interface} at the same decision of \code{insert\_tubes}. Notes are shortened.},label={lst:calls}]
move_eef(targets={left_x: -0.19, left_y: 0.025, left_z: 0.88},
  note="Approach the left tube near its capped end, leaving
        its tapered end free for insertion.")
move_eef_relative(targets={left_dx: 0.115, left_dy: 0.22,
  left_dz: -0.065}, note="Approach the left tube from above ...")
approach(arm="left", target_object_id="tube_left",
  target_pose={position: {x: -0.2, y: 0.015, z: 0.91}},
  note="Translate the now downward-facing open hand ...")
move_eef(targets={left_x: -0.19, left_y: -0.015, left_z: 0.94},
  num_actions=32, note="Approach above the left tube ...")
move_eef_sequence(num_moves=2, moves=[
  {targets: {left_pitch_deg: 45}, note="Turn the open hand ..."},
  {targets: {left_pitch_deg: 0, left_x: -0.2, left_y: 0.015,
             left_z: 0.87}, note="Approach above the capped end ..."}])
\end{lstlisting}

\begin{figure}[!ht]
\centering
\includegraphics{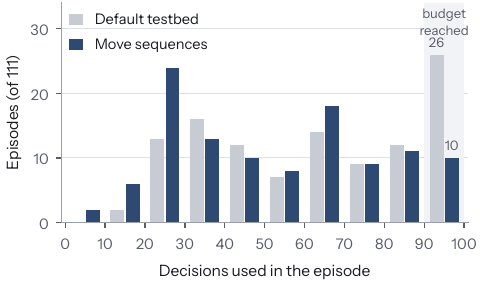}
\caption{\textbf{Episodes by decisions used} on the evaluation set, default testbed against move sequences. The shaded bin holds the episodes that reach the budget of 100 decisions.}
\label{fig:moveseq}
\end{figure}

\paragraph{Move sequences.}
Allowing several moves per decision shortens episodes, and the six episodes that used to run out of the 100-decision budget account for most of the gain (\cref{tab:moveseq,fig:moveseq}).
If they keep their old scores, the gain over the default largely disappears, which is why \justrua lifts the budget instead.

\begin{table}[!ht]
\centering
\caption{\textbf{Move sequences} against the default testbed, 111 episodes each with a budget of 100 decisions. The capped six are the episodes that run out of decisions under the default.}
\label{tab:moveseq}
\footnotesize
\setlength{\tabcolsep}{3pt}
\begin{tabular}{lcc}
\toprule
& Default & Sequences \\
\midrule
Evaluation set & 38.0 & 42.0 \\
\quad capped six keep old scores & \na & 39.4 \\
Decisions per episode (mean) & 63.2 & 52.7 \\
Decisions, typical episode & 65 & 51 \\
Decisions, 31 both finish & 50.2 & 40.0 \\
Successes, capped six & 0 & 3 \\
Mean score, capped six & 11.7 & 58.3 \\
\bottomrule
\end{tabular}
\end{table}

\paragraph{Use of each interface.}
With semantic operations the model used the named operations in most of its calls and \code{adjust\_view} throughout the first round.
With move sequences it sent more than one move in a sizeable share of its decisions.
With a model-chosen step count its choices split between short motions of a few steps and long ones near the upper limit.
With the self-built library of \cref{sec:loop}, library operations made up a small share of its calls and the rest went to \code{move\_eef}.

\section{Evaluation Details}
\label{app:eval}

\paragraph{Tasks run on four layouts.}
The evaluation set runs 23 tasks on four layouts each.
On each of them a partial result scores at most 50, and the default testbed both succeeded and failed on it within eight rounds, at low or medium effort.
\begin{itemize}
\item \emph{Generalization}: push\_T, sort\_nesting\_dolls\_by\_size, stack\_bowls, fold\_clothes, arrange\_largest\_number, pack\_objects\_into\_box.
\item \emph{Memory}: swap\_T, match\_and\_pick\_from\_conveyor, imitate\_sorting\_sequence, press\_by\_number, cover\_blocks.
\item \emph{Open}: align\_blocks, solve\_equation, stack\_blocks\_by\_language, classify\_objects\_by\_language, general\_pickup, pick\_from\_conveyor\_by\_image.
\item \emph{Precision}: plug\_in\_charger, fasten\_screws, play\_Xylophone, pour\_balls\_into\_vase.
\item \emph{Long-horizon}: put\_bottles\_into\_dustbin, classify\_objects.
\end{itemize}

\paragraph{How much the evaluation set helps.}
On the eight rounds used to choose them, the 23 tasks account for about 95\% of the task-level variance in the overall score.
At medium effort the bootstrap standard deviation of the overall score is 4.8 points for a single round and 2.6 for the evaluation set, close to the 2.4 of running every task four times with 168 episodes.
The bootstrap resamples each task's rounds independently, so it measures sensitivity to the layouts in our history.

\paragraph{Fixed layouts.}
The official evaluation scores each task over 50 episodes, since the score of a task on any single layout is spiky.
We run far fewer episodes, so which layouts we run matters.
Each round runs the next layout of every task in a fixed order, and rounds are not equally hard.
Every configuration in this paper therefore runs on the same fixed layouts, in the first round and on the evaluation set alike, rather than on fresh random draws.

\paragraph{Layout order and weighting.}
Every round uses seed 0 and the next layout of each task in a fixed order.
For the twelve Generalization tasks, which have both standard and random layouts, rounds alternate between the two, starting with random layout 0, so the Generalization group of the evaluation set weights random layouts at 75\% against 50\% in the official evaluation.
We report evaluation-set and full scores separately for this reason.
The 23 four-layout tasks were chosen from the history of the default testbed, which does not guarantee that the evaluation set ranks every new design as the full benchmark would.

\begin{figure}[!ht]
\centering
\includegraphics{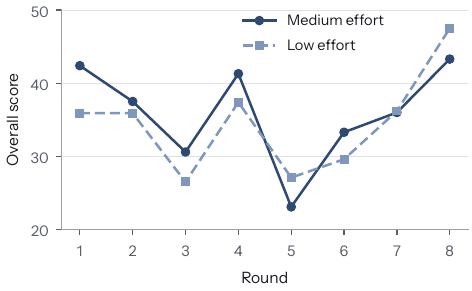}
\caption{\textbf{Overall score of the default testbed} in each of eight rounds at low and medium effort. Both settings see the same layouts in each round.}
\label{fig:rounds}
\end{figure}

\paragraph{Round-to-round variation.}
Low and medium effort rise and fall together across eight rounds, with a correlation of 0.84, although we have not separated layout difficulty from run-to-run randomness (\cref{fig:rounds}).
A score from the first round alone tends to run high.
In all three settings of \cref{tab:rounds} the first round scored above the mean of eight rounds, and for the official harness the eight-round mean is close to the full evaluation.
\Cref{tab:first-round} gives the first-round score of each design \justrua keeps, the rings of \cref{fig:teaser}, next to its evaluation-set score.

\begin{table}[!ht]
\centering
\begin{minipage}[t]{0.48\linewidth}
\centering
\caption{\textbf{First round against the mean of eight rounds.} Official L3 rounds map our eight layouts onto the episodes published by RoboProbe \citep{zhang2026roboprobe}.}
\label{tab:rounds}
\footnotesize
\setlength{\tabcolsep}{3pt}
\begin{tabular}{@{}lccc@{}}
\toprule
& \makecell{First\\round} & \makecell{Mean of\\8 rounds} & \makecell{Full\\evaluation} \\
\midrule
Official L3 harness & 32.3 & 28.7 & 29.0 \\
Default, low effort & 35.9 & 34.5 & \na \\
Default, medium effort & 42.4 & 35.9 & \na \\
\bottomrule
\end{tabular}
\end{minipage}\hfill
\begin{minipage}[t]{0.48\linewidth}
\centering
\caption{\textbf{Scores of \code{swap\_T}} on the default testbed in each of eight rounds.}
\label{tab:swapT}
\footnotesize
\setlength{\tabcolsep}{3pt}
\begin{tabular}{@{}lcccccccc@{}}
\toprule
Round & 1 & 2 & 3 & 4 & 5 & 6 & 7 & 8 \\
\midrule
Low effort & 0 & 100 & 0 & 0 & 0 & 100 & 100 & 100 \\
Medium effort & 0 & 0 & 100 & 100 & 0 & 100 & 0 & 100 \\
\bottomrule
\end{tabular}
\end{minipage}
\end{table}

\begin{table}[!ht]
\centering
\caption{\textbf{First round against the evaluation set} for the designs \justrua keeps, in the order added. The first round is the 42-episode subset of the evaluation set that runs the first layout of each task.}
\label{tab:first-round}
\small
\setlength{\tabcolsep}{6pt}
\begin{tabular}{@{}lcc@{}}
\toprule
Configuration & First round (42 episodes) & Evaluation set (111 episodes) \\
\midrule
Official L3 harness & 32.3 & 29.7 \\
Codex, full history & 43.5 & 34.7 \\
20-step cap (default testbed) & 42.4 & 38.0 \\
Compact robot readings & 38.3 & 39.8 \\
Predicted depth & 47.7 & 42.4 \\
Coordinate solving (\justrua) & 54.0 & 50.3 \\
\bottomrule
\end{tabular}
\end{table}

\paragraph{An uncertain task.}
\code{swap\_T} is a typical case.
It scores 100 only when the two T blocks have swapped places within 20\,mm, each at the required orientation within 3$^\circ$, and both arms are back at rest, and anything less scores 0.
Both settings scored 0 in the first round, and each succeeded in four of the eight rounds (\cref{tab:swapT}).

\section{Behaviour Review}
\label{app:review}

Each first-round episode is reviewed by its own GPT-6.1 Sol agent at medium effort, which takes no part in running the robot.
The agent reads the complete action log, including the note the model writes with each call, and the videos of the three cameras.
It scans each view at short intervals, checks consecutive frames around every executed action, every suspicious moment and the start and end of each event, and records each event with its time range, the views that show it and the frames used as evidence.
Events that span several actions, such as repeated empty grasps ending in a regrasp or a handover in several stages, are linked into one record.
A main agent then checks the coverage of each review and every episode's counts against this evidence and the native logs, and sends episodes that do not match back for another review.
Success and score come only from the simulator.
\Cref{tab:behaviour-defs} gives the definitions and \cref{tab:behaviour-all} the counts for every reviewed configuration.

\begin{table}[!ht]
\centering
\caption{\textbf{Behaviour definitions} used in the review.}
\label{tab:behaviour-defs}
\footnotesize
\renewcommand{\arraystretch}{1.12}
\begin{tabularx}{\linewidth}{@{}>{\raggedright\arraybackslash}p{0.17\linewidth} >{\raggedright\arraybackslash}X >{\raggedright\arraybackslash}X@{}}
\toprule
Behaviour & Counts & Does not count \\
\midrule
Empty grasp & The gripper closes on an object but does not hold it. Each separate attempt counts once, whether or not it is retried. & Closing on purpose to press or push, or closing an empty hand. \\
Dropped object & An object that has been lifted and held is lost before the model releases it. & An object that tips over or lands off target after a deliberate release. \\
Fine placement failure & One attempt at insertion, hanging, assembly, stacking or precise alignment that does not meet the task's requirement. Backing off and trying again starts a new attempt. & Small corrections within one continuous attempt, and ordinary drops during transport. \\
View adjustment & An executed motion made to see something, by the acting arm or by the free arm, which may move to look at an occluded area or out of the way. Counted once per motion phase. & Normal approach motions, and moves made only to avoid a collision. \\
Handover & An executed transfer of one object between the arms, directly from gripper to gripper or through the table or another support. & A transfer planned only in text, or picking up an object the other arm dropped by accident. \\
Unwanted contact & Real, unintended contact between the arms, or between an arm and an object. One continuous contact counts once. & Intended contact in grasping, pushing, pressing or assembly, and motions the planner rejected. \\
\bottomrule
\end{tabularx}
\end{table}

\begin{table}[!ht]
\centering
\caption{\textbf{Behaviour counts} over the same 42 first-round episodes for every reviewed configuration.
Free-arm views are the view adjustments made by the free arm, and handovers are split into direct ones and those through the table or another support.
Counts are lower bounds, and a behaviour that could not be checked in an episode is left out rather than counted as zero.
Coordinate solving runs without a decision cap, so its first row counts the events that start within the first 100 decisions, which makes it comparable with the other rows, and its second row counts whole episodes.
Shaded rows are kept in \justrua.}
\label{tab:behaviour-all}
\footnotesize
\setlength{\tabcolsep}{2.0pt}
\begin{tabular}{@{}l ccc cc cc cc@{}}
\toprule
& & & & \multicolumn{2}{c}{View adjustments} & \multicolumn{2}{c}{Handovers} & \multicolumn{2}{c}{Unwanted contact} \\
\cmidrule(lr){5-6}\cmidrule(lr){7-8}\cmidrule(lr){9-10}
Configuration & \makecell{Empty\\grasps} & Drops & \makecell{Fine\\placement} & All & \makecell{Free\\arm} & Direct & \makecell{Via a\\support} & \makecell{Arms} & \makecell{Arm,\\object} \\
\midrule
\grouprow{9}{Harness and action}
\kept Default testbed & 47 & 26 & 20 & 59 & 34 & 5 & 10 & 6 & 7 \\
Low effort & 34 & 21 & 26 & 48 & 31 & 3 & 11 & 0 & 2 \\
High effort & 42 & 27 & 22 & 64 & 50 & 4 & 11 & 1 & 4 \\
Relative targets & 54 & 30 & 25 & 54 & 34 & 5 & 10 & 6 & 20 \\
Semantic operations & 47 & 35 & 19 & 211 & 93 & 8 & 10 & 5 & 12 \\
Move sequences & 38 & 27 & 20 & 45 & 29 & 3 & 11 & 0 & 1 \\
Model-chosen steps & 45 & 29 & 22 & 61 & 35 & 5 & 9 & 1 & 11 \\
\midrule
\grouprow{9}{Reusing experience}
Visual demonstration & 50 & 20 & 16 & 42 & 20 & 2 & 16 & 0 & 3 \\
Text trajectory & 53 & 17 & 25 & 66 & 28 & 8 & 14 & 0 & 6 \\
Self-built library & 33 & 22 & 28 & 71 & 37 & 4 & 9 & 0 & 5 \\
\midrule
\grouprow{9}{Scene geometry}
Full JSON readings & 47 & 33 & 23 & 55 & 32 & 3 & 8 & 4 & 11 \\
\kept Compact readings & 41 & 31 & 18 & 56 & 32 & 2 & 11 & 5 & 16 \\
Simulator depth & 17 & 33 & 15 & 48 & 25 & 2 & 7 & 2 & 8 \\
\kept Predicted depth & 16 & 28 & 7 & 49 & 31 & 4 & 11 & 2 & 13 \\
Drawn, sim. depth & 17 & 35 & 15 & 51 & 29 & 8 & 6 & 4 & 12 \\
Drawn, RGB and depth, sim. & 25 & 31 & 27 & 45 & 27 & 6 & 5 & 7 & 5 \\
Drawn with depth, pred. & 39 & 30 & 14 & 62 & 39 & 4 & 5 & 4 & 9 \\
Affordance panel, pred. & 34 & 32 & 16 & 51 & 40 & 7 & 7 & 5 & 11 \\
\kept Coordinate solving & 15 & 34 & 7 & 41 & 28 & 6 & 5 & 4 & 8 \\
\quad whole episodes & 20 & 37 & 10 & 52 & 30 & 6 & 8 & 4 & 9 \\
\bottomrule
\end{tabular}
\end{table}

\section{Demonstrations}
\label{app:demos}

\begin{figure}[!ht]
\centering
\includegraphics[width=0.72\linewidth]{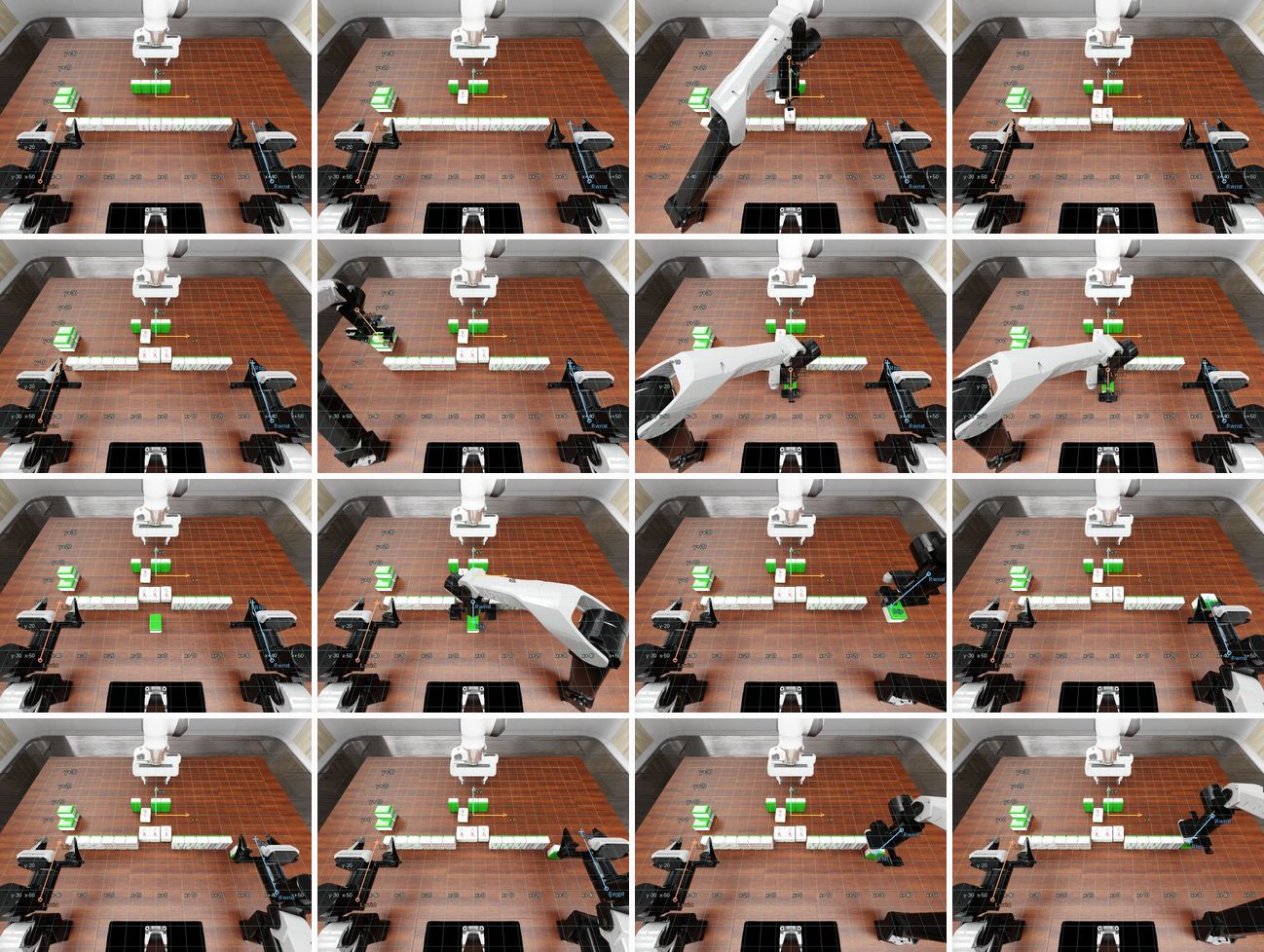}
\caption{\textbf{Visual demonstration} for \code{make\_kong}, the 16 keyframes of the source episode, head camera, in reading order.
The model sees them once before its first decision.
Three tasks also include the author's wrist-camera views.
No action trace or robot state is attached.}
\label{fig:keyframes}
\end{figure}

\Cref{fig:keyframes} shows what the model sees before its first decision with the visual demonstration.
The text trajectory gives the same episode in about 16,000 characters per task on average, injected once and kept in the conversation.
\Cref{lst:trajectory} covers three of its 40 stages for \code{make\_kong}.
The handover appears only as coordinates, the left hand carrying the tile to the table centre and opening there, and the right hand closing at the same point.

\begin{lstlisting}[float=ht,caption={\textbf{Text trajectory}, three of the 40 stages of the \code{make\_kong} demonstration.},label={lst:trajectory}]
Stage 27 (source turn 31, control step 1642):
left: pose=[-0.001,-0.23,0.7789,0.0,0.0,90.0]; opening=0.78
right: pose=[0.3,-0.2019,0.922,90.0,0.0,0.0]; opening=0.18
Source event: LEFT arm closed on an object with the fingertips
  at z = 81.3 (still holding it).
1. left gripper 1.00 => left_gripper=1; result=ok
Stage 29 (source turn 33, control step 1892):
left: pose=[-0.3,-0.2019,0.922,90.0,0.0,0.0]; opening=1
right: pose=[0.3,-0.2019,0.922,90.0,0.0,0.0]; opening=0.18
1. right gripper 1.00 => right_gripper=1; result=ok
2. right point down => nominal right target_pose=
   [0.3,-0.207,0.9169,0.0,0.0,0.0]; result=ok
3. right move x -20.0 => nominal right target_pose=
   [0.1,-0.207,0.9169,0.0,0.0,0.0]; result=ok
4. right move x -10.1 => nominal right target_pose=
   [-0.001,-0.207,0.9169,0.0,0.0,0.0]; result=ok
Stage 30 (source turn 36, control step 2052):
left: pose=[-0.3,-0.2019,0.922,90.0,0.0,0.0]; opening=1
right: pose=[-0.001,-0.208,0.7759,0.0,0.0,0.0]; opening=1
1. right gripper 0.00 => right_gripper=0; result=ok
\end{lstlisting}

\section{Self-Improvement Loop}
\label{app:loop}

\Cref{tab:loop} lists each round of the loop in \cref{sec:loop}.
The agent's changes follow the failures it saw, and most of them make an operation stop earlier so the model can check the result before going on.

\begin{table}[!ht]
\centering
\caption{\textbf{Five rounds of the self-improvement loop.} The first library, written from the demonstrations, holds \code{grasp}, \code{transport}, \code{place} and \code{push}. Scores are per task, from 0 to 1. Each round runs on a different group, so the group results are not a learning curve. On the tasks the loop ran, the runs inside the loop average 35.4 and the round-5 library 36.4. Observations and changes paraphrase the library's own records.}\label{tab:loop}
\footnotesize
\setlength{\tabcolsep}{3.5pt}
\renewcommand{\_}{\textunderscore\allowbreak}
\begin{tabularx}{\linewidth}{@{}c >{\raggedright\arraybackslash\hsize=0.95\hsize}X >{\raggedright\arraybackslash\hsize=0.95\hsize}X >{\raggedright\arraybackslash\hsize=1.1\hsize}X@{}}
\toprule
Round & Group and per-task scores & Seen in the runs & Changed for the next round \\
\midrule
1 & stack\_bowls 1, plug\_in\_charger 0, general\_pickup 1, fill\_pen\_holder 0, sweep\_blocks 0, deposit\_coin 0.2, cover\_blocks 0, classify\_objects 0.4, push\_T 1 &
In plug\_in\_charger and fill\_pen\_holder the model let go before the plug or the pen was in place. &
\code{place} and \code{grasp} can stop before letting go or closing, so the model can look first. A new \code{insert} lines a held tip up with a hole and pushes it in, only once the hand is within 2\,mm and 1$^\circ$ of the entry. \\
\midrule
2 & hang\_mugs 0.15, insert\_tubes 0.2, align\_blocks 0, put\_bottles\_into\_dustbin 0.25, stack\_blocks 1, insert\_key 0.15, pour\_balls\_into\_vase 0, build\_tower 0.1, pack\_objects\_into\_box 0.1 &
Rising straight up after letting go pulled the tube into a lean in insert\_tubes and caught the handle in hang\_mugs. In pour\_balls\_into\_vase repeated turns moved the lip off the vase and spilled the balls. &
\code{place} can back away in a chosen direction before rising. A new \code{pivot} turns a held object while one point on it, such as a lip or a tip, stays where it is. \\
\midrule
3 & sort\_nesting\_dolls\_by\_size 0, fill\_egg\_holder 0.1, fold\_clothes 1, play\_Xylophone 0, swap\_blocks 0 &
In play\_Xylophone and swap\_blocks the model counted presses that never registered. &
The guide tells the model to record an attempted press separately from a confirmed one. A new \code{tap} makes one press and returns, only once the hand is within 2\,mm and 1$^\circ$ of the target. \\
\midrule
4 & make\_toast 0, press\_by\_number 1, make\_kong 0, arrange\_largest\_number 1, stack\_blocks\_by\_language 1, classify\_objects\_by\_language 1, store\_tools\_in\_toolbox 0 &
In arrange\_largest\_number and stack\_blocks\_by\_language long carries failed, and the model succeeded only after breaking them into shorter moves itself. In store\_tools\_in\_toolbox the tool slipped as it was lifted. &
\code{transport} can split a carry into short segments, and can stop right after the lift so the model can check the grip. \\
\midrule
5 & store\_laptop\_and\_headphones 0, match\_and\_pick\_from\_conveyor 1, pour\_liquid\_into\_cup 0, play\_tic\_tac\_toe 0.1, pick\_from\_conveyor\_by\_image 1, pour\_by\_language 0 &
In play\_tic\_tac\_toe the model waited for the opponent one short step at a time and used up its decisions. &
A new \code{hold} keeps both arms and jaws still for a set number of steps, so a long wait takes one decision. \\
\bottomrule
\end{tabularx}
\end{table}

\section{Scene Geometry Details}
\label{app:geometry}

\paragraph{Robot readings.}
\Cref{lst:readings} shows the compact format of \justrua for a left gripper that holds an object and an open right gripper.
Full JSON gives the same values with vectors, thresholds and identifiers for every finger and takes about 3,100 characters per decision.

\begin{lstlisting}[float=ht,caption={\textbf{Compact robot readings} at one decision.},label={lst:readings}]
O1 sensors:
left_finger_position_mm=[0, 0]
left_finger_velocity_mm_s=[-1.13, 0.02]
left_drive_target_mm=[-10, -10]
left_finger_joint_force_N=[-12.22, -12.11]
left_normal_N=[12.13, 12.12]; contact_ge1N=[yes, yes]
left_wrist_total_force_N=0.68
left_wrist_total_torque_Nm=0.029
right_finger_position_mm=[44, 44]
right_finger_velocity_mm_s=[0.01, 0.01]
right_drive_target_mm=[44, 44]
right_finger_joint_force_N=[0, 0]
right_normal_N=[0, 0]; contact_ge1N=[no, no]
right_wrist_total_force_N=0
right_wrist_total_torque_Nm=0
\end{lstlisting}

\paragraph{Depth images.}
Depth from either source is drawn as camera-forward distance in metres on a fixed colour scale from 0 to 2\,m, one image per camera, labelled with the camera it belongs to.
With depth the text of each observation also gives the pose, intrinsics and resolution of the three cameras, and a few more gripper readings, namely the gap between the fingers, the velocity of the grasp point, how long the current contact has lasted with its recent peak force, and the contact points.
For predicted depth, Depth Pro runs on each RGB image as it is captured.
It receives no measured focal length, camera parameters or simulator depth, its output is not rescaled, and a failed prediction is reported as unavailable while the RGB image is kept.
The instruction states that predicted depth is an estimate, that it is least reliable around thin objects, occlusion boundaries and unfamiliar scenes, and that it does not show contact or millimetre-level geometry.

\begin{figure}[!ht]
\centering
\includegraphics[width=\linewidth]{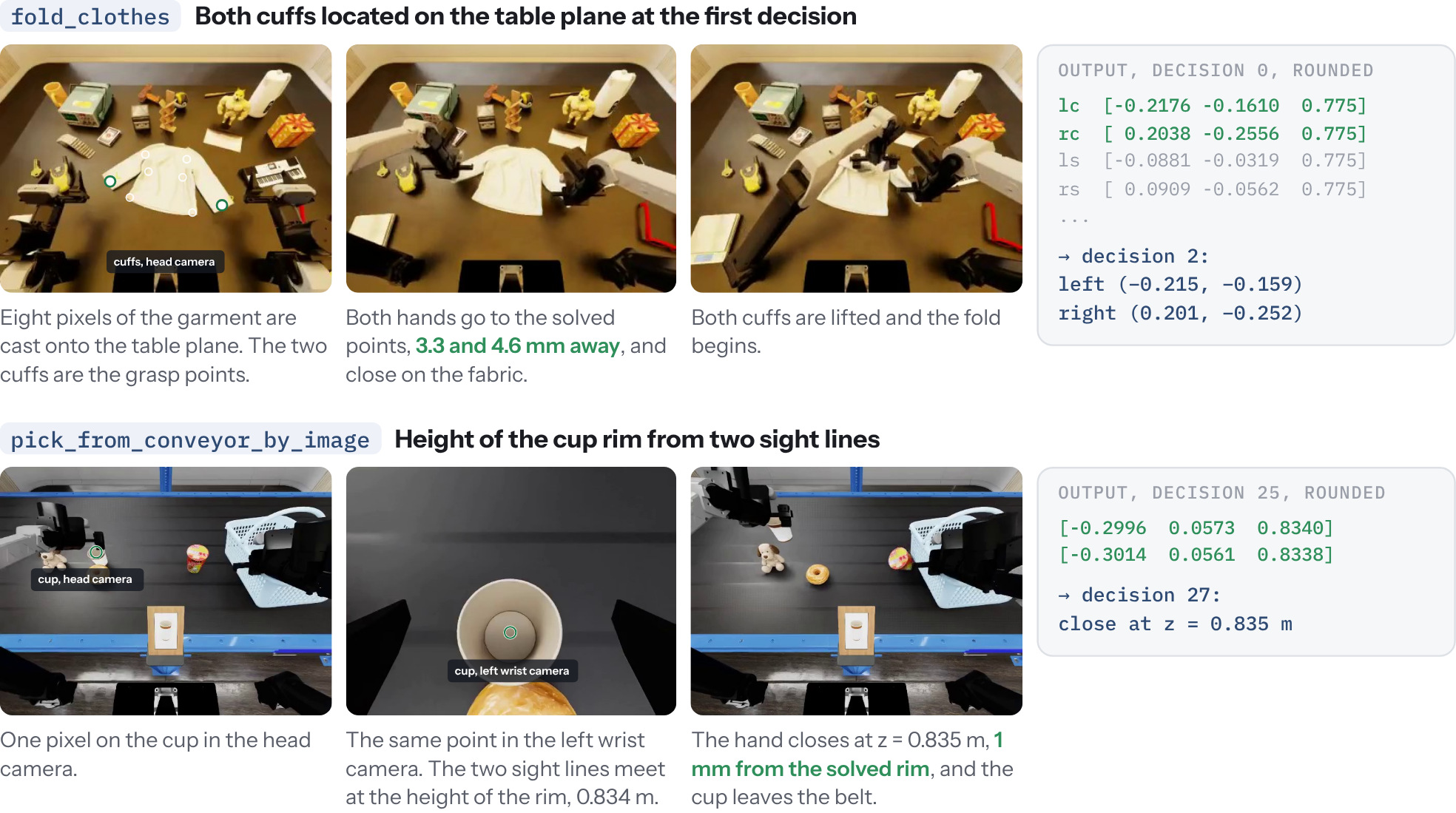}
\caption{\textbf{More coordinate solving.}
Top, \code{fold\_clothes}. The model locates both cuffs on the table plane, and both hands then close within a few millimetres of the solved points.
Bottom, \code{pick\_from\_conveyor\_by\_image}. Two sight lines to the cup, from the head and the left wrist camera, meet at a height of 0.834\,m, and the hand closes 1\,mm from the solved rim.
Distances compare the output of the code with the commands and gripper positions that followed, not with the true positions of the objects.}
\label{fig:solver-more}
\end{figure}

\paragraph{Coordinate solving.}
\Cref{lst:code} is the first code the model runs in \code{insert\_tubes}.
It casts rays from five pixels through the head camera and meets the table plane at $z=0.782$\,m.
At its next decision the model sends the left gripper to within 1.2\,mm of the tube axis.
Later in the episode it triangulates the holes of the rack from the head and right wrist cameras (\cref{fig:coord}).
\Cref{fig:solver-more} shows the same use of code in two other tasks.

\begin{lstlisting}[float=ht,caption={\textbf{First code run} in \code{insert\_tubes}, with the first lines of its output.},label={lst:code}]
python3 - <<'PY'
import numpy as np
for u,v in [(220,202),(201,235),(215,212),(302,177),(319,186)]:
 d=np.array([(u-320)/288.1325,-.86603*(v-240)/288.1325+.5,
             -.5*(v-240)/288.1325-.86603])
 p=np.array([0,-.41,1.308]); q=p+d*((.782-p[2])/d[2]); print(u,v,q)
PY
220 202 [-0.22816849 -0.00619792  0.782     ]
201 235 [-0.2533848  -0.09402191  0.782     ]
215 212 [-0.23449104 -0.03411066  0.782     ]
...
\end{lstlisting}

\section{Error Analysis Details}
\label{app:errors}

\Cref{tab:error-tasks} lists every task behind \cref{tab:errors}.

\begin{table}[!ht]
\centering
\caption{\textbf{Failures by task.} Every first-round task where \justrua scores below 50, sorted by cause. A task can appear under more than one cause.}\label{tab:error-tasks}
\footnotesize
\begin{tabularx}{\linewidth}{@{}l l >{\raggedright\arraybackslash}X@{}}
\toprule
Cause & Task & What goes wrong \\
\midrule
\multirow{8}{*}{Strategy}
& make\_kong & One arm cannot flip the tile and place it at once. The model never hands it over. \\
& deposit\_coin & Tries a sideways handover and the coin drops. Keeping it in one hand works. \\
& pour\_balls\_into\_vase & Pours forward, which the arm does poorly. The task needs a side pour. \\
& pour\_by\_language & Pours forward and the grip slips. \\
& pour\_liquid\_into\_cup & Pours forward and blocks its own view. \\
& pick\_from\_conveyor\_by\_image & Grasps the basket at a poor point and cannot time the belt. \\
& sweep\_blocks & Leaves the receiving arm an awkward grasp point in the handover. \\
& hang\_mugs & Changes view to compensate, but still cannot predict where the mug handle lands. \\
\midrule
\multirow{6}{*}{Collision}
& pack\_objects\_into\_box & Knocks the light box out of place. \\
& make\_toast & Misjudges the size of the bread, so neither slice goes in. \\
& build\_tower & One knock brings the whole tower down. \\
& fill\_egg\_holder & A knock shuts the egg box. \\
& fill\_pen\_holder & Knocks waste steps it needs later. \\
& put\_bottles\_into\_dustbin & Earlier knocks pile up and leave no room to recover. \\
\midrule
\multirow{6}{*}{Unstable grasp}
& sort\_nesting\_dolls\_by\_size & Moves too fast with a round doll, which slips. \\
& insert\_key & The key's odd friction makes it slip in the handover the task requires. \\
& deposit\_coin & The coin has few stable grasp points and drops in the handover. \\
& sweep\_blocks & The irregular blocks behave oddly in simulation. \\
& store\_tools\_in\_toolbox & Too many irregular tools to handle. \\
& store\_laptop\_and\_headphones & The task itself is very hard. \\
\midrule
\multirow{5}{*}{Prompt or scoring}
& fasten\_screws & May be unable to tell whether it met the required tolerance. \\
& play\_Xylophone & A simple task with no visible failure. \\
& swap\_blocks & Looks simple, yet every run fails. \\
& pack\_objects\_into\_box & Looks complete but scores low. The box seems to have moved. \\
& pour\_liquid\_into\_cup & Part of the failure looks like a prompt or scoring problem. \\
\bottomrule
\end{tabularx}
\end{table}

\paragraph{Prompt revision.}
For the five tasks marked as prompt or scoring failures, we rewrote the task recipe so that it states the conditions the scorer checks, such as tolerances, counts and the final pose, and reran the first-round layout of each (\cref{lst:prompt-fix}).
This raises the first-round score only slightly (\cref{tab:errors}).
The revision draws on the scorer of each test task, so we report it as part of the error analysis and do not count it in \justrua.

\begin{lstlisting}[caption={\textbf{Revised recipe} for \code{fasten\_screws}, requirements part. The original recipe gives only the description and the scoring ladder.},label={lst:prompt-fix}]
Goal: Complete all three same-color screw-and-nut assemblies. Each nut must be
upright, centered on its matching screw and seated to the required depth; then
release all parts and return both arms to their starting poses.

## Task requirements
Match each screw with the nut of the same color. Placing matching parts
together or making a turning motion is not sufficient: each pair must end in
the required assembled geometry.
For a completed pair, the nut must be less than 10 degrees from upright. The
horizontal alignment tolerance for a matched pair is strictly less than
0.001 m. The highest point of the screw must be more than 0.009 m above the
lowest point of the nut. ...
Keep completed pairs assembled while working on the others. Partial-credit
checks require both grippers to be at least 80% open. ...
\end{lstlisting}

\end{document}